\documentclass[lettersize,journal]{IEEEtran}
\usepackage{amsmath,amsfonts}
\usepackage{algorithmic}
\usepackage{algorithm}
\usepackage{array}
\usepackage{textcomp}
\usepackage{stfloats}
\usepackage{url}
\usepackage{verbatim}
\usepackage{graphicx}
\usepackage{cite}
\usepackage{booktabs}
\usepackage{multirow}
\usepackage{rotating}
\usepackage{wrapfig}
\usepackage{xcolor}
\usepackage{colortbl}
\usepackage{amssymb}
\usepackage{subcaption}

\definecolor{eccvblue}{rgb}{0.12,0.49,0.85}
\definecolor{nbarrier}{RGB}{255, 180, 75}
\definecolor{nbicycle}{RGB}{255, 192, 203}
\definecolor{nbus}{RGB}{255, 255,130}
\definecolor{ncar}{RGB}{150, 200, 245}
\definecolor{npedestrian}{RGB}{255, 220, 200}
\definecolor{nterrain}{RGB}{180, 240, 150}
\definecolor{grey}{RGB}{230,230,230}
\definecolor{purple}{RGB}{200, 200, 245}
\definecolor{cyan}{RGB}{184,246,228}

\usepackage[colorlinks,linkcolor=red]{hyperref}
\usepackage{listings}

\begin{document}

\title{DD-RobustBench: An Adversarial Robustness Benchmark for Dataset Distillation}

\author{Yifan Wu, Jiawei Du$^{\star}$, Ping Liu$^{\star}$, Yuewei Lin, Wei Xu$^{\dagger}$ and Wenqing Cheng
\thanks{Yifan Wu, Wei Xu and Wenqing Cheng are with Huazhong University of Science and Technology, China (email: \{yifanwu, xuwei, chengwq\}@hust.edu.cn)}
\thanks{Jiawei Du is with Centre for Frontier AI Research (CFAR), IHPC, A*STAR, Singapore (email: dujw@cfar.a-star.edu.sg)}
\thanks{Ping Liu is with University of Nevada, Reno, NV, USA (email: pino.pingliu@gmail.com)}
\thanks{Yuewei Lin is with Brookhaven National Laboratory, Upton, NY, USA (email: ywlin@bnl.gov)}
\thanks{$^{\star}$ means authors contributed equally to this work. $^{\dagger}$ means the corresponding author.}
}

\markboth{Journal of \LaTeX\ Class Files,~Vol.~14, No.~8, August~2021}%
{Shell \MakeLowercase{\textit{et al.}}: A Sample Article Using IEEEtran.cls for IEEE Journals}

\maketitle

\begin{abstract}
Dataset distillation techniques have revolutionized the way of utilizing large datasets by compressing them into smaller, yet highly effective subsets that preserve the original datasets' accuracy. 
However, while these methods have proven effective in reducing data size and training times, the robustness of these distilled datasets against adversarial attacks remains underexplored. 
This vulnerability poses significant risks, particularly in security-sensitive applications.
To address this critical gap, we introduce DD-RobustBench, a novel and comprehensive benchmark specifically designed to evaluate the adversarial robustness of distilled datasets. 
Our benchmark is the most extensive of its kind and integrates a variety of dataset distillation techniques, including recent advancements such as TESLA, DREAM, SRe2L, and D4M, which have shown promise in enhancing model performance.
DD-RobustBench also rigorously tests these datasets against a diverse array of adversarial attack methods to ensure broad applicability. 
Our evaluations cover a wide spectrum of datasets, including but not limited to, the widely used ImageNet-1K. This allows us to assess the robustness of distilled datasets in scenarios mirroring real-world applications.
Furthermore, our detailed quantitative analysis investigates how different components involved in the distillation process, such as data augmentation, downsampling, and clustering, affect dataset robustness. 
Our findings provide critical insights into which techniques enhance or weaken the resilience of distilled datasets against adversarial threats, offering valuable guidelines for developing more robust distillation methods in the future.
Through DD-RobustBench, we aim not only to benchmark but also to push the boundaries of dataset distillation research by highlighting areas for improvement and suggesting pathways for future innovations in creating datasets that are not only compact and efficient but also secure and resilient to adversarial challenges.
The implementation details and essential instructions are available on \href{https://github.com/FredWU-HUST/DD-RobustBench}{DD-RobustBench}. 
\end{abstract}


\section{Introduction}
\label{introduction}
\IEEEPARstart{D}{atasets} play a pivotal role in the field of machine learning. 
The last decade has witnessed an unprecedented increase in the size and quality of datasets across various domains such as image recognition, natural language processing, and autonomous driving \cite{russakovsky2015imagenet,lin2014coco,everingham2015pascal}. 
These datasets not only facilitate the development of more sophisticated algorithms but also challenge existing computational paradigms by requiring substantial storage capacities and prolonging the duration of model training \cite{Ge:relation,Wu:pruning,Liao:grounding,Zhao:cotraining}.
This escalation in dataset scale poses significant logistical and environmental costs, highlighting a critical need for more efficient data handling techniques.

In response to these challenges, dataset distillation, or condensation, emerges as a potent solution. 
This technique aims to compress voluminous datasets into smaller, yet highly effective sets \cite{wang:DD,Zhao:DC,Zhao:DSA,Zhao:DM,Cazenavette:MTT}. 
By doing so, it preserves the essential characteristics necessary for maintaining competitive performance across various application fields \cite{Such:nas,Sangermano:continue1,Xiong:federated}. 
Distillation not only mitigates the issues related to storage and computational demands but also expedites the training process, thus enhancing the overall efficiency and sustainability.

Previous studies on dataset distillation have primarily focused on optimizing the distillation process to maintain model accuracy with significantly reduced dataset sizes. 
These studies, targeting specific compression ratios, have predominantly aimed to demonstrate that distilled datasets can indeed support effective model training, emphasizing efficiency without a corresponding loss in performance \cite{Zhao:DC,Zhao:DSA,Zhao:DM,Cazenavette:MTT,cui2023scaling}. 
However, this focus on accuracy overlooks other critical aspects of model performance, notably adversarial robustness, which is a pivotal measure of model resilience in trustworthy applications. 
Conversely, the body of research dedicated to adversarial robustness has been extensively developed for original datasets \cite{goodfellow2014explaining,madry2017towards,carlini2017towards,croce2020minimally}. 
This research has established robust frameworks for understanding and improving the resistance of models trained on comprehensive datasets, but has not yet been systematically applied to their distilled counterparts. 
The lack of focus on distilled datasets within this research area suggests a gap in understanding how distillation techniques might affect or possibly even compromise the adversarial robustness of models.

Addressing this gap, we identify two critical areas that warrant further exploration: the impact of training on distilled data on model robustness relative to training on original datasets, and the effects of specific components used in the distillation process on adversarial robustness. 
This exploration is essential not only for understanding the limits and potentials of distilled datasets in adversarial settings, but also for advancing dataset distillation techniques to ensure that they do not inadvertently decrease model robustness.

In this study, we introduce DD-RobustBench, a comprehensive benchmark to assess the adversarial robustness of distilled datasets. 
Our benchmark establishes a unified standard for evaluating recent distillation methods across various settings and hyperparameters, expanding upon prior benchmarks \cite{2023arXiv230503355C}. 
We conducted extensive evaluations using prominent distillation methods such as TESLA \cite{cui2023scaling}, SRe2L \cite{yin2023squeeze}, and D4M \cite{su2024d4m}, and performed comparative analyses on large-scale datasets like ImageNet-1K\cite{russakovsky2015imagenet} and smaller datasets such as CIFAR-10/100 \cite{krizhevsky2009learning}.
Our findings suggest that distilled datasets generally exhibit greater robustness than their original counterparts, with smaller distilled sets often outperforming larger ones.
Further, we explored how various distillation components, such as data augmentation\cite{Zhao:DSA}, downsampling factors\cite{kim2022idc}, and clustering numbers\cite{Liu:DREAM}, affect robustness. 
Our quantitative experiments reveal that while clustering has minimal impact, certain augmentations and a higher downsampling fraction detrimentally affect robustness. 
These insights aim to guide and inspire future research in dataset distillation.

The contribution of our work can be summarized as follows:
\begin{itemize}
    \item We propose a comprehensive and unified benchmarking pipeline for evaluating the adversarial robustness of distilled datasets. This framework facilitates extensive comparisons across three key dimensions: dataset varieties, distillation methods, and adversarial attacks. Detailed instructions on how to utilize our benchmark are available in the supplementary materials.

    \item Our experiments demonstrate that models trained on distilled datasets exhibit superior robustness compared to those trained on original datasets, with smaller distilled datasets achieving better robust accuracy.

    \item We investigated the impact of different components used in distillation process on robustness, including data augmentation, downsampling factor and clustering number. Quantitative results indicate that clustering has minor impact on robustness while augmentations like crop and cutout as well as too large fraction factor would cause robustness drop.
\end{itemize}
\vspace{-1em}
\section{Related Work}
\subsection{Dataset Distillation} 
The objective of dataset distillation is to produce a dataset that is significantly smaller in scale yet maintains competitive performance compared to the original dataset. 
This method has been extensively applied across various machine learning fields including neural architecture search \cite{Such:nas,Burnaev:nas}, continual learning \cite{Sangermano:continue1,Rosasco:continue}, and federated learning \cite{Xiong:federated,Liu:federated}, demonstrating its versatility and importance.

The concept was initially formulated as a bi-level optimization problem by Wang et al. \cite{wang:DD}. 
Building on this, Nguyen et al. \cite{Nguyen:KIP} introduced Kernel Inducing Points (KIP) for kernel regression, which marked a significant advance by providing closed-form solutions. 
Following studies diversified the focus to encompass not just the final model output but also intermediary training dynamics such as gradients \cite{Zhao:DC,Zhao:DSA}, feature distributions \cite{Zhao:DM,Wang:CAFE,zhao2023improved,son2024fyi,zhang2024dance}, and training trajectories \cite{Cazenavette:MTT, Guo:Lossless,liu2024att,yang2024nsd}.
Recent advancements have also targeted the efficiency of distillation techniques, adjusting elements like labels \cite{Bohdal:label,Sucholutsky:label} and sample selection \cite{Liu:DREAM,Liu:DREAM+,xu2023distill}. 
The intrinsic correlation within distilled data has been another area of focus, leading to improved understandings of data interdependencies \cite{kim2022idc,du2023sequential,deng2024inter,shul2024one}.

With the challenge of scaling distillation to large datasets such as ImageNet-1K, recent efforts \cite{yin2023squeeze,cui2023scaling,yin2023dataset} have introduced novel strategies such as optimization decoupling \cite{yin2023squeeze}, unrolled gradient computation \cite{cui2023scaling}, and curriculum data augmentation \cite{yin2023dataset}. 
In addition, innovative approaches have incorporated generative models such as GANs and diffusion models into the distillation process, using their generative capabilities to reduce resource demands \cite{gu2024efficient,su2024d4m,li2024generative,su2024diffusion}. 
The robustness of distilled datasets has also been enhanced through methods like regularization during the squeezing stage in SRe2L \cite{xuel2024curvature,yin2023squeeze}.

Despite significant progress, the robustness of distilled datasets, particularly their ability to withstand adversarial conditions, remains largely unexplored. 
This gap underscores a critical area for future research, suggesting that while dataset distillation can enhance efficiency and performance, its implications for model security and reliability require further investigation.

\vspace{-0.75em}
\subsection{Adversarial Robustness}
\label{:sec:advrobust}
Since the revelation of vulnerabilities in deep neural networks to adversarial examples, there has been significant effort in developing increasingly potent adversarial attack methods. 
Szegedy et al. \cite{szegedy2013intriguing} initially treated the generation of adversarial examples as a box-constrained optimization problem. 
Building on this, Goodfellow et al. \cite{goodfellow2014explaining} introduced the Fast Gradient Sign Method (FGSM), a foundational one-step attack that spurred various derivatives such as the Basic Iterative Method (BIM) \cite{kurakin2016adversarial} and the momentum iterative FGSM \cite{dong2018boosting}. 
Additionally, the Projected Gradient Descent (PGD) \cite{madry2017towards} and the CW attack \cite{carlini2017towards} have become benchmarks for evaluating the robustness of deep models.
Further contributions by Moosavi-Dezfooli et al. \cite{moosavi2016deepfool} quantified adversarial robustness by computing the minimal adversarial perturbation. 
Croce et al. \cite{AA_atack_2020} later introduced AutoAttack, a per-sample attack comprising four diverse methods: APGD$_{CE}$, APGD$^T_{DLR}$, targeted FAB \cite{croce2020minimally}, and Square Attack \cite{andriushchenko2020square}. 
Recent innovations have considered gradient variance in iterative methods \cite{wang2021vmifgsm} and have introduced novel loss functions \cite{schwinn2023jitter} to enhance attack effectiveness.
The ongoing development of these sophisticated attack techniques underscores the critical need for robust adversarial defenses, particularly in applications involving dataset distillation. 
\vspace{-0.75em}
\subsection{Benchmarks on Dataset Distillation}
With the proliferation of dataset distillation methods, there is a growing need to establish a standardized framework for evaluating and comparing these techniques effectively. 
Cui et al. \cite{Cui:dcbench} introduced DC-Bench, the first benchmark designed to facilitate comparisons of dataset distillation methods \cite{Cui:dcbench}. 
While DC-Bench has effectively addressed standard accuracy, it has not yet extended its scope to include adversarial robustness, which is a critical aspect of model performance under adversarial conditions \cite{hu2023understanding}.

In response to evolving benchmark needs, Chen et al. \cite{2023arXiv230503355C} proposed a framework that not only considers privacy and fairness but also begins to address the robustness of distilled data. 
This framework, while pioneering, primarily examines model robustness under specific conditions such as the DeepFool Attack \cite{moosavi2016deepfool}. 
It has not yet incorporated other prevalent adversarial methods such as the Fast Gradient Sign Method (FGSM) \cite{goodfellow2014explaining}, Projected Gradient Descent (PGD) \cite{madry2017towards}, CW attack \cite{carlini2017towards}, VMIFGSM \cite{wang2021vmifgsm}, Jitter \cite{schwinn2023jitter}, and AutoAttack \cite{AA_atack_2020}, which are crucial for a thorough evaluation of adversarial robustness in distilled data. 
Additionally, the framework's limited public availability may pose challenges for widespread usage and validation by the research community.

Given these considerations, there is a clear opportunity to further develop benchmarks that not only build upon existing efforts but also expand their scope to include a broader range of adversarial conditions. 
Our work seeks to fill this gap by creating a more inclusive and publicly accessible benchmark that addresses the full spectrum of adversarial threats, thereby enhancing the security and reliability of dataset distillation techniques.

\begin{figure*}[!t]
  \centering
  \includegraphics[width=1\linewidth]{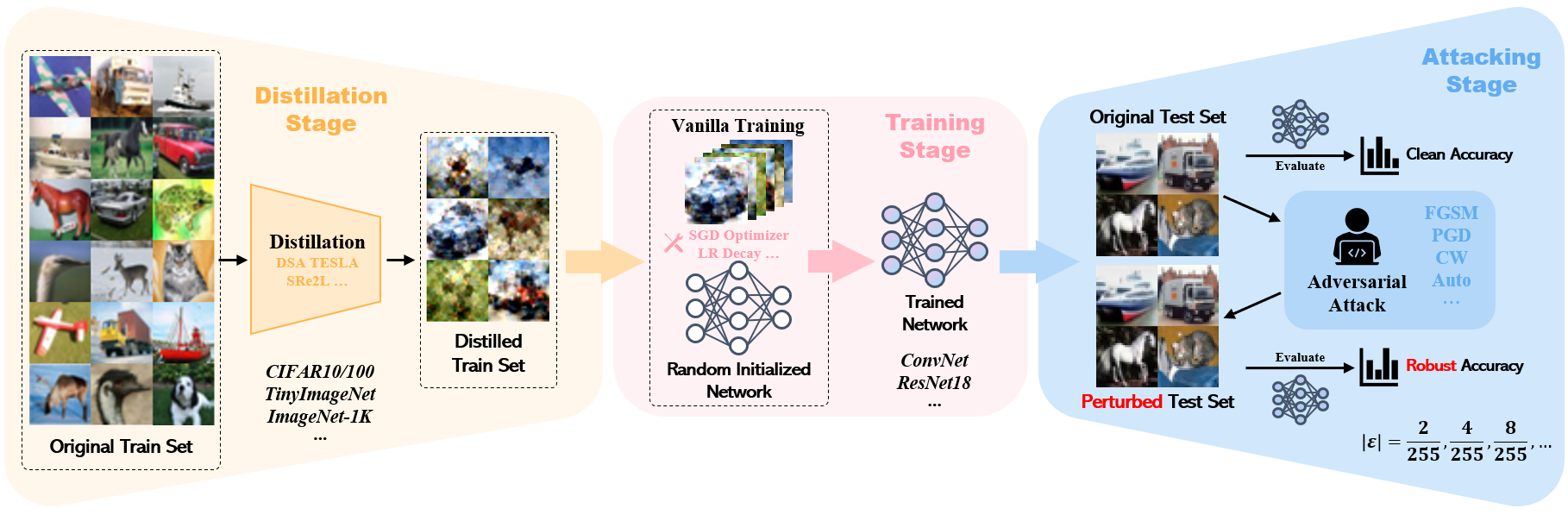}
  \caption{Illustration of our robustness evaluation pipeline. We first utilize dataset distillation for condensing the train set. Subsequently, the distilled data is used to train a random initialized network from scratch. The trained network is then evaluated on the perturbed original test set which is manipulated by adversarial attacks.
  }
  \label{fig:pipeline2}
\end{figure*}
\vspace{-1em}
\section{Our Investigations}
\label{sec:Investigations}
In this section, we begin by outlining the objectives of the experiments as detailed in Section \ref{sec:Overview}. 
Subsequently, in Section \ref{sec:setup}, we elaborate on the experimental setup and provide a thorough overview of the datasets, models, and attack algorithms.
In Section \ref{sec:implement} we provide other pertinent implementation details used in our study.
Following this, we present our experimental findings and endeavor to extract discernible patterns and conclusions in Section \ref{sec:results}. In Section \ref{sec:component}, we show a qualitative and quantitative analysis on the impact of different distillation components on robustness.
\vspace{-0.75em}
\subsection{Experimental Overview}
\label{sec:Overview}
Our research focuses on the analysis and examination of adversarial robustness. 
To be specific, we aim to determine whether models trained on distilled data retain their accuracy on adversarially perturbed test sets.
To assess the impact of distilled data on model robustness, our study conducts a comprehensive comparative analysis across various configurations, including images per class (IPC), different datasets, and distillation algorithms. We have designed a unified evaluation pipeline to ensure a fair comparison of these diverse data sources.

Figure \ref{fig:pipeline2} illustrates our robustness evaluation pipeline of distilled data. 
Initially, we utilize a distillation algorithm to distill the original train set, yielding a condensed tiny dataset. 
Subsequently, we train a randomly initialized model using these distilled data, in line with previous studies  \cite{Cui:dcbench} \cite{Zhao:DC} \cite{Cazenavette:MTT} primarily concerned with accuracy assessment. 
The trained model and the original test set are then subjected to various attacking algorithms, which introduce perturbations to the original test samples. 
The performance of the model is evaluated on the basis of its robust accuracy against this perturbed test set, providing insights into the defensive capabilities of distilled datasets under adversarial conditions.

\begin{table*}[p] 
  \caption{Clean and robust accuracies of models trained on distilled CIFAR-10, CIFAR-100 and TinyImageNet. The perturbation budget is set to $|\varepsilon|=2/255$.}
\label{table:benchmark2/255}
\setlength{\tabcolsep}{2.5pt} 
\tiny
\renewcommand{\arraystretch}{1} 
\centering
\resizebox{\textwidth}{!}{
\begin{tabular}{cc|ccccc|ccccc|ccccc}
\toprule
\multicolumn{2}{c|}{Dataset} &\multicolumn{5}{c|}{\textbf{CIFAR-10}} & \multicolumn{5}{c|}{\textbf{CIFAR-100}} & \multicolumn{5}{c}{\textbf{TinyImageNet}} \\
\hline
\multicolumn{2}{c|}{IPC}   & 1 & 5 & 10 & 30 & 50 & 1 & 5 & 10 & 30 & 50 & 1 & 5 & 10 & 30 & 50 \\ \specialrule{0.8pt}{0pt}{0pt}
\rowcolor{nbarrier}\multirow{6}{*}{\rotatebox{90}{\textcolor{nbarrier}{$\blacksquare$} DC}} & \textbf{Clean} & \textbf{29.73} & \textbf{41.23} & \textbf{46.07} & \textbf{53.65} & \textbf{55.08} & \textbf{12.82} & \textbf{21.97} & \textbf{25.66} & \textbf{30.82} & \textbf{30.19} & \textbf{5.49} & \textbf{9.70} & \textbf{11.74} & \textbf{13.51} & \textbf{11.97} \\
 & FGSM & 20.12 &20.74 &19.84 &19.87 &17.71 &6.32 &10.34 &11.53 &8.41 &6.31 &1.26 &2.18 &1.85 &0.65 &0.57\\
 & PGD & 19.78 & 19.68 & 18.27 & 17.55 & 14.82 & 6.08 & 9.67 & 10.77 & 7.13 & 4.87 & 1.12 & 1.90 & 1.46 & 0.33 & 0.34 \\
 & CW & 18.41 & 18.75 & 16.50 & 16.62 & 14.23 & 4.82 & 8.61 & 10.07 & 6.37 & 4.38 & 0.55 & 1.35 & 0.92 & 0.26 & 0.35 \\
 & VMI &19.73 & 19.69 & 18.20 & 17.55 & 14.73 & 6.08 & 9.61 & 10.75 & 7.14 & 4.87 & 1.10 & 1.90 & 1.45 & 0.33 & 0.33 \\
 & Jitter & 18.49 & 19.09 & 17.20 & 17.89 & 15.63 & 4.84 & 8.67 & 10.05 & 6.57 & 4.52 & 0.57 & 1.37 & 0.94 & 0.21 & 0.35 \\\hline
\rowcolor{nbicycle}\multirow{6}{*}{\rotatebox{90}{\textcolor{nbicycle}{$\blacksquare$} DSA}} & \textbf{Clean} & \textbf{29.27} & \textbf{48.20} & \textbf{52.93} & \textbf{56.41} & \textbf{61.14} & \textbf{14.38} & \textbf{27.13} & \textbf{32.94} & \textbf{37.29} & \textbf{42.87} & \textbf{5.47} & \textbf{13.99} & \textbf{17.47} & \textbf{17.12} & \textbf{21.89} \\
 & FGSM & 17.24 &21.74 &22.04 &16.40 &16.97 &7.12 &11.16 &11.45 &7.56 &10.81 &1.62 &2.91 &3.38 &2.02 &1.67 \\
 & PGD & 16.82 & 20.14 & 20.01 & 13.85 & 14.81 & 6.76 & 9.91 & 9.72 & 5.54 & 8.46 & 1.49 & 2.30 & 2.62 & 1.40 & 1.00 \\
 & CW & 16.63 & 19.76 & 19.83 & 13.55 & 13.84 & 6.05 & 9.36 & 9.54 & 5.44 & 8.53 & 1.06 & 2.01 & 2.37 & 1.09 & 0.99 \\
 & VMI & 16.84 & 20.19 & 20.06 & 13.83 & 13.93 & 6.75 & 9.93 & 9.68 & 5.40 & 8.30 & 1.45 & 2.29 & 2.63 & 1.41 & 1.01 \\
 & Jitter & 16.64 & 20.29 & 20.70 & 14.74 & 15.41 & 6.11 & 9.44 & 9.66 & 5.85 & 8.65 & 1.03 & 2.00 & 2.36 & 1.11 & 0.99 \\\hline
\rowcolor{nbus}\multirow{6}{*}{\rotatebox{90}{\textcolor{nbus}{$\blacksquare$} DM}} & \textbf{Clean} & \textbf{26.75} & \textbf{42.78} & \textbf{49.81} & \textbf{59.97} & \textbf{63.12} & \textbf{11.71} & \textbf{23.90} & \textbf{29.92} & \textbf{38.48} & \textbf{43.56} & \textbf{4.06} & \textbf{9.76} & \textbf{14.16} & \textbf{21.03} & \textbf{21.29} \\
 & FGSM & 18.38 &15.88 &17.07 &22.96 &22.83 &6.12 &7.47 &8.59 &8.54 &8.39 &1.35 &1.27 &1.83 &1.87 &1.48 \\
 & PGD & 18.20 & 14.58 & 14.74 & 20.17 & 19.36 & 5.83 & 6.36 & 6.74 & 6.36 & 6.00 & 1.18 & 0.95 & 1.27 & 1.10 & 0.74 \\
 & CW & 18.01 & 14.68 & 14.95 & 20.23 & 19.47 & 5.34 & 6.17 & 6.81 & 6.58 & 6.02 & 0.96 & 0.87 & 1.28 & 1.05 & 0.63 \\
 & VMI & 18.23 & 14.71 & 14.87 & 20.03 & 19.15 & 5.79 & 6.47 & 6.71 & 6.09 & 5.87 & 1.21 & 0.98 & 1.29 & 1.07 & 0.75 \\
 & Jitter & 18.04 & 15.27 & 16.01 & 21.19 & 20.83 & 5.32 & 6.08 & 6.86 & 6.56 & 6.29 & 0.99 & 0.87 & 1.23 & 1.09 & 0.66 \\\hline
\rowcolor{ncar}\multirow{6}{*}{\rotatebox{90}{\textcolor{ncar}{$\blacksquare$} MTT}} & \textbf{Clean} & \textbf{45.74} & \textbf{57.19} & \textbf{60.98} & \textbf{66.23} & \textbf{70.45} & \textbf{20.30} & \textbf{34.81} & \textbf{37.84} & \textbf{42.65} & \textbf{44.34} & \textbf{8.91} & \textbf{14.86} & \textbf{19.93} & \textbf{23.05} & \textbf{26.16} \\
 & FGSM & 21.27 &21.57 &24.12 &22.30 &23.48 &8.62 &10.81 &10.13 &8.98 &10.15 &1.67 &1.63 &1.79 &1.22 &0.90 \\
 & PGD & 19.79 & 19.45 & 21.95 & 18.82 & 19.19 & 8.27 & 9.17 & 8.03 & 6.59 & 7.29 & 1.45 & 1.25 & 1.20 & 0.67 & 0.41 \\
 & CW & 16.83 & 17.99 & 19.00 & 18.29 & 18.87 & 6.08 & 7.73 & 7.18 & 6.06 & 6.94 & 0.84 & 0.64 & 0.79 & 0.47 & 0.36 \\
 & VMI & 19.65 & 19.37 & 21.57 & 18.79 & 18.92 & 8.23 & 9.08 & 7.90 & 6.44 & 7.03 & 1.40 & 1.21 & 1.17 & 0.71 & 0.42 \\
 & Jitter & 17.65 & 19.01 & 20.19 & 20.42 & 20.84 & 6.23 & 7.81 & 7.40 & 6.45 & 7.13 & 0.82 & 0.64 & 0.80 & 0.49 & 0.40 \\\hline
\rowcolor{npedestrian}\multirow{6}{*}{\rotatebox{90}{\textcolor{npedestrian}{$\blacksquare$} TESLA}} & \textbf{Clean} & \textbf{47.07} & \textbf{57.23} & \textbf{61.78} & \textbf{67.08} & \textbf{68.83} & \textbf{20.00} & \textbf{31.65} & \textbf{34.99} & \textbf{39.16} & \textbf{45.92} & \textbf{7.40} & \textbf{16.01} & \textbf{19.45} & \textbf{23.71} & \textbf{26.63} \\
 & FGSM & 20.25 &21.30 &20.04 &23.28 &19.28 &7.90 &10.18 &6.44 &7.81 &10.16 &1.18 &1.15 &1.35 &1.38 &0.92 \\
 & PGD & 19.10 & 19.06 & 17.11 & 19.81 & 15.62 & 7.39 & 8.66 & 4.78 & 5.32 & 7.20 & 1.02 & 0.89 & 0.86 & 0.80 & 0.46 \\
 & CW & 17.93 & 17.77 & 16.10 & 19.18 & 15.32 & 5.67 & 7.62 & 3.82 & 4.96 & 6.81 & 0.56 & 0.51 & 0.42 & 0.61 & 0.37 \\
 & VMI & 19.12 & 19.06 & 17.11 & 19.74 & 15.59 & 7.38 & 8.55 & 4.37 & 5.27 & 7.16 & 1.02 & 0.88 & 0.83 & 0.82 & 0.44 \\
 & Jitter & 18.44 & 19.21 & 17.78 & 20.84 & 17.49 & 5.69 & 7.64 & 4.05 & 5.20 & 7.15 & 0.55 & 0.50 & 0.45 & 0.62 & 0.34 \\\hline
\rowcolor{nterrain}\multirow{6}{*}{\rotatebox{90}{\textcolor{nterrain}{$\blacksquare$} SRe2L}} & \textbf{Clean} & \textbf{13.49} & \textbf{31.06} & \textbf{37.53} & \textbf{54.88} & \textbf{63.28} & \textbf{4.68} & \textbf{28.92} & \textbf{39.64} & \textbf{51.33} & \textbf{53.95} & \textbf{6.28} & \textbf{18.38} & \textbf{26.92} & \textbf{39.49} & \textbf{43.24} \\
 & FGSM & 7.67 &15.09 &14.94 &18.89 &21.39 &2.39 &8.47 &10.96 &14.17 &15.08 &1.73 &1.75 &2.69 &4.77 &5.71 \\
 & PGD & 7.17 & 13.95 & 13.09 & 15.29 & 16.12 & 2.35 & 6.85 & 7.08 & 7.69 & 8.04 & 1.50 & 1.11 & 1.59 & 2.40 & 2.70 \\
 & CW & 6.17 & 13.77 & 12.92 & 15.53 & 16.76 & 1.62 & 4.46 & 4.97 & 6.26 & 6.27 & 1.26 & 1.03 & 1.67 & 2.70 & 2.94 \\
 & VMI & 6.96 & 14.09 & 13.28 & 15.39 & 15.73 & 2.34 & 5.72 & 5.48 & 5.62 & 5.49 & 1.52 & 1.11 & 1.54 & 2.40 & 2.60 \\
 & Jitter & 6.22 & 14.40 & 14.07 & 17.38 & 19.49 & 1.56 & 4.66 & 5.93 & 7.67 & 8.29 & 1.23 & 0.88 & 1.43 & 2.35 & 2.72 \\\hline
 \rowcolor{grey}\multirow{6}{*}{\rotatebox{90}{\textcolor{grey}{$\blacksquare$} IDM}} & \textbf{Clean} & \textbf{46.14} & \textbf{54.45} & \textbf{58.84} & \textbf{65.83} & \textbf{67.82} & \textbf{23.90} & \textbf{40.02} & \textbf{45.33} & \textbf{45.83} & \textbf{46.13} & \textbf{10.42} & \textbf{22.39} & \textbf{22.4} & \textbf{24.96} & \textbf{26.89} \\
 & FGSM & 22.79 &28.52 &25.36 &27.25 &25.66 &8.43 &11.71 &11.69 &9.58 &7.88 &2.44 &2.95 &1.97 &1.56 &2.74 \\
 & PGD & 21.55 & 26.79 & 23.34 & 24.18 & 22.47 & 7.27 & 9.34 & 9.17 & 6.61 & 5.11 & 2.17 & 2.17 & 1.15 & 0.91 & 1.69 \\
 & CW & 21.26 & 26.41 & 23.17 & 24.39 & 22.74 & 6.88 & 9.46 & 9.31 & 6.83 & 5.31 & 1.92 & 2.21 & 1.22 & 0.95 & 1.65 \\
 & VMI & 21.60 & 26.85 & 23.33 & 24.07 & 22.51 & 7.31 & 9.34 & 8.96 & 6.31 & 4.91 & 2.19 & 2.20 & 1.16 & 0.86 & 1.71 \\
 & Jitter & 21.71 & 26.99 & 24.21 & 25.87 & 24.60 & 6.78 & 9.47 & 9.37 & 6.96 & 5.57 & 2.00 & 2.20 & 1.17 & 0.99 & 1.63 \\\hline
 \rowcolor{purple}\multirow{6}{*}{\rotatebox{90}{\textcolor{purple}{$\blacksquare$} DREAM}} & \textbf{Clean} & \textbf{44.66}&\textbf{57.07}&\textbf{63.25}&\textbf{67.75}&\textbf{68.96}&\textbf{24.95} &\textbf{36.17} &\textbf{42.36} &\textbf{44.46} &\textbf{46.84} &\textbf{7.59} &\textbf{17.58} &\textbf{18.88} &\textbf{19.91} &\textbf{17.46} \\
 & FGSM & 23.31 &23.45 &24.08 &21.79 &23.10 &9.48 &8.17 &9.36 &6.99 &6.78 &1.05 &1.82 &1.18 &0.71 &1.65 \\
 & PGD & 22.13 &20.99 &20.98 &17.90 &19.29 &8.49 &6.26 &6.99 &4.47 &4.46 &0.88 &1.21 &0.70 &0.26 &1.14 \\
 & CW & 21.61 &20.81 &20.98 &18.08 &19.54 &7.83 &6.20 &7.07 &4.56 &4.54 &0.59 &1.14 &0.68 &0.28 &0.71 \\
 & VMI & 22.32 &21.13 &20.87 &18.02 &19.20 &8.48 &6.27 &6.98 &4.41 &4.33 &0.91 &1.26 &0.66 &0.28 &1.10 \\
 & Jitter & 22.07 &21.86 &22.54 &19.92 &21.53 &7.93 &6.28 &7.28 &4.90 &4.95 &0.62 &1.16 &0.68 &0.29 &0.80 \\\hline

 \rowcolor{cyan}\multirow{6}{*}{\rotatebox{90}{\textcolor{cyan}{$\blacksquare$} D4M}} & \textbf{Clean} & \textbf{23.39}&\textbf{42.34}&\textbf{48.16}&\textbf{63.53}&\textbf{69.81}&\textbf{11.27} &\textbf{31.90} &\textbf{40.12} &\textbf{48.15} &\textbf{51.10} &\textbf{2.16} &\textbf{6.94} &\textbf{14.25} &\textbf{35.80} &\textbf{42.89} \\
 & FGSM & 9.67 &17.19 &22.07 &25.72 &27.13 &3.37 &5.91 &6.33 &6.40 &5.17 &0.63 &1.37 &1.81 &3.92 &6.30 \\
 & PGD & 8.86 &15.47 &20.14 &22.20 &23.08 &2.84 &4.18 &4.25 &3.97 &3.14 &0.53 &1.05 &0.97 &1.91 &3.14 \\
 & CW & 8.35 &15.38 &20.16 &22.62 &23.71 &2.84 &4.61 &4.68 &4.28 &3.46 &0.50 &0.98 &1.12 &2.29 &3.51 \\
 & VMI & 8.92 &15.50 &20.14 &22.33 &23.22 &2.86 &4.21 &4.14 &3.79 &2.92 &0.52 &1.04 &0.96 &1.70 &2.85 \\
 & Jitter & 8.24 &15.20 &19.85 &22.95 &24.58 &3.36 &5.17 &5.19 &5.00 &3.80 &0.48 &0.90 &1.06 &1.97 &3.10 \\\hline


\multicolumn{2}{c|}{\multirow{4}{*}{\textbf{Whole}}} 
& &\textbf{Clean}&\textbf{FGSM}&\textbf{PGD}& & &\textbf{Clean}&\textbf{FGSM}&\textbf{PGD}& & &\textbf{Clean}&\textbf{FGSM}&\textbf{PGD}& \\& 
& &84.45&21.26&14.92& & &55.74&3.07&3.98& & &38.77&4.92&3.20& \\&

& &\textbf{CW}&\textbf{VMI}&\textbf{Jitter}& & &\textbf{CW}&\textbf{VMI}&\textbf{Jitter}& & &\textbf{CW}&\textbf{VMI}&\textbf{Jitter}& \\& 
& & 15.34 & 14.19 &18.87& & & 1.18 & 0.86 &1.80& & & 2.53 & 3.00 &2.62& \\
 \bottomrule
\end{tabular}
}
\end{table*}

\begin{figure*}[t]
    \centering
    \includegraphics[width=\linewidth]{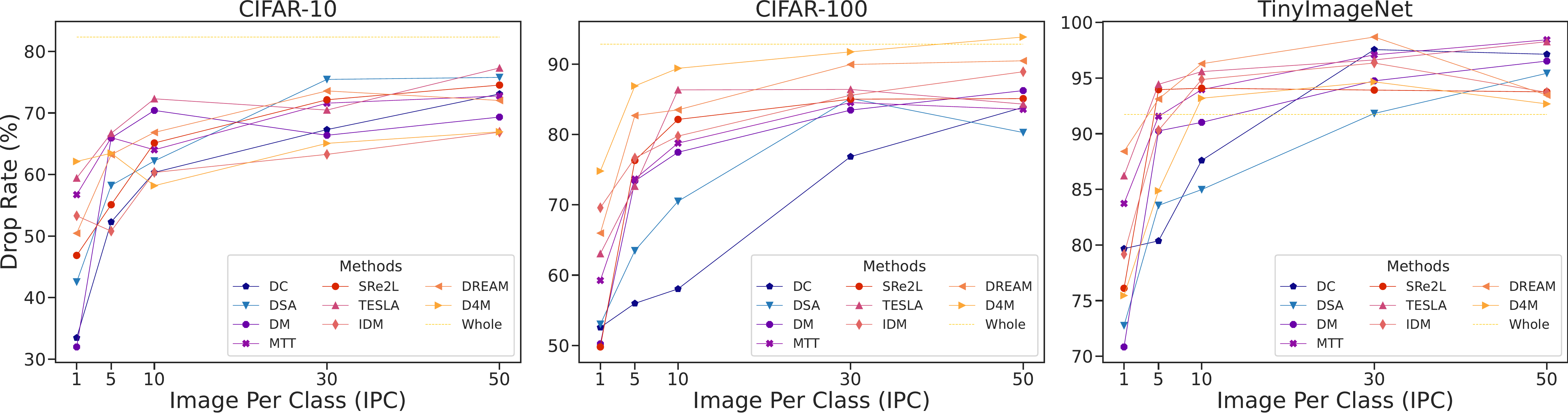}
    \caption{Drop rates after PGD attack. The dashed lines represent the $DR$ based on the original dataset. Higher $DR$ means relatively worse robustness.}
    \label{fig:droprate}
\end{figure*}

\begin{table*}[t]
  \caption{Original accuracies and robust accuracies of models trained on ImageNet-subsets. MTT is applied in the distillation stage. The perturbation budget is set to $\left| \varepsilon \right|=2/255$.}\label{table:imagenetsubset2/255}
\centering
\setlength{\tabcolsep}{2.5pt} 
\renewcommand{\arraystretch}{1.1} 
\resizebox{1\textwidth}{!}{
\begin{tabular}{c|cccccc|cccccc|cccccc}
\toprule
      \multirow{2}{*}{IPC}               & \multicolumn{6}{c|}{\cellcolor{nbarrier}\textbf{ImageNette}}        & \multicolumn{6}{c|}{\cellcolor{nbicycle}\textbf{ImageWoof}}         & \multicolumn{6}{c}{\cellcolor{nbus}\textbf{ImageFruit}}        \\
 & \textbf{Clean} & FGSM  & PGD   & CW& VMI&Jitter    & \textbf{Clean} & FGSM  & PGD   & CW  & VMI&Jitter  & \textbf{Clean} & FGSM  & PGD   & CW  & VMI&Jitter   \\
 \hline
1 & \textbf{48.20} & 19.40 & 18.60 & 17.60 & 18.60 & 17.80 & \textbf{30.40} & 6.20 & 5.20 & 3.60 & 5.60 & 4.20 & \textbf{25.00} & 6.80 & 6.00 & 5.40 & 6.00 & 6.00 \\
5 & \textbf{62.00} & 28.40 & 24.60 & 23.80 & 24.80 & 25.00 & \textbf{35.20} & 4.80 & 3.60 & 2.80 & 3.80 & 3.40 & \textbf{41.00} & 7.40 & 6.00 & 5.00 & 5.80 & 5.80 \\
10 & \textbf{66.40} & 27.80 & 23.60 & 22.60 & 23.80 & 23.80 & \textbf{38.00} & 5.80 & 4.00 & 3.40 & 4.20 & 4.60 & \textbf{42.20} & 11.20 & 9.20 & 9.00 & 9.00 & 10.00 \\
30 & \textbf{66.60} & 25.00 & 20.40 & 19.80 & 20.60 & 23.00 & \textbf{38.80} & 3.20 & 2.20 & 2.40 & 2.40 & 2.80 & \textbf{44.40} & 9.20 & 6.00 & 5.20 & 6.00 & 6.40 \\
50 & \textbf{67.60} & 25.20 & 20.60 & 20.60 & 19.80 & 23.40 & \textbf{39.40} & 3.40 & 1.80 & 1.60 & 1.80 & 2.20 & \textbf{44.60} & 9.80 & 7.20 & 7.60 & 7.20 & 9.00 \\
\rowcolor{grey!40}\textbf{Whole} & \textbf{86.40} & 32.80 & 20.40 & 20.80 & 20.00 & 24.00 & \textbf{67.20} & 2.00 & 0.60 & 1.00 & 0.60 & 2.60 & \textbf{67.40} & 10.40 & 5.60 & 5.80 & 5.80 & 8.80   \\ \toprule
      \multirow{2}{*}{IPC}                 & \multicolumn{6}{c|}{\cellcolor{ncar}\textbf{ImageMeow}}         & \multicolumn{6}{c|}{\cellcolor{npedestrian}\textbf{ImageSquawk}}       & \multicolumn{6}{c}{\cellcolor{nterrain}\textbf{ImageYellow}}       \\ 
 & \textbf{Clean} & FGSM  & PGD   & CW & VMI&Jitter  & \textbf{Clean} & FGSM  & PGD   & CW & VMI&Jitter     & \textbf{Clean} & FGSM  & PGD   & CW  & VMI&Jitter    \\

\hline
1 & \textbf{31.00} & 5.40 & 5.00 & 4.40 & 5.00 & 4.60 & \textbf{39.00} & 11.80 & 10.60 & 10.20 & 10.80 & 11.00 & \textbf{44.60} & 17.20 & 15.80 & 14.40 & 15.80 & 15.40 \\
5 & \textbf{41.40} & 7.00 & 4.80 & 4.40 & 4.80 & 6.60 & \textbf{52.40} & 18.60 & 15.40 & 14.40 & 15.20 & 16.20 & \textbf{59.20} & 24.60 & 21.20 & 20.20 & 21.20 & 21.20 \\
10 & \textbf{44.40} & 8.00 & 5.60 & 5.20 & 5.60 & 6.40 & \textbf{55.60} & 16.20 & 12.40 & 12.00 & 12.40 & 14.00 & \textbf{63.40} & 23.40 & 18.80 & 18.20 & 19.00 & 21.00 \\
30 & \textbf{44.20} & 5.40 & 3.80 & 4.00 & 4.20 & 4.40 & \textbf{56.20} & 12.40 & 10.00 & 10.20 & 10.00 & 11.80 & \textbf{65.40} & 21.20 & 18.00 & 18.00 & 18.00 & 19.80 \\
50 & \textbf{44.20} & 4.00 & 2.80 & 3.00 & 2.80 & 4.60 & \textbf{59.20} & 13.60 & 11.00 & 10.60 & 11.60 & 13.60 & \textbf{66.20} & 21.20 & 17.40 & 17.20 & 17.40 & 20.20 \\
\rowcolor{grey!40}\textbf{Whole} & \textbf{69.00} & 5.00 & 1.80 & 1.60 & 1.60 & 6.80 & \textbf{86.40} & 29.00 & 19.80 & 20.40 & 19.20 & 24.80 & \textbf{84.40} & 29.00 & 20.40 & 20.60 & 19.40 & 25.00 \\ \bottomrule
\end{tabular}}
\end{table*}

\begin{table}
\caption{Robust accuracy of ImageNet-1K obtained by SRe2L. The perturbation budget is set to $\left| \varepsilon \right|=2/255$.
  }\label{table:imagenet1k}
\centering
\setlength{\tabcolsep}{2.5pt} 
\tiny
\renewcommand{\arraystretch}{1.1} 
\resizebox{0.485\textwidth}{!}{
\begin{tabular}{cccccccc}
\hline
\multirow{2}{*}{IPC} & \multicolumn{7}{c}{\cellcolor{grey!40}\textbf{ImageNet-1K}} \\
 & \textit{\textbf{Clean}} & FGSM & PGD & CW & VMI & Jit & Auto \\ \hline
1 & \textbf{1.98} & 0.09 & 0.05 & 0.04  & 0.04 & 0.05 & 0.01 \\
5 & \textbf{21.63} & 0.16 & 0.04 & 0.04 & 0.03 & 0.06 & 0.01 \\
10 & \textbf{31.49} & 0.23 & 0.04 & 0.04 & 0.03 & 0.12 & 0.02 \\
30 & \textbf{42.98} & 0.49 & 0.07 & 0.05 & 0.02 & 0.10 & 0.01 \\
50 & \textbf{46.13} & 0.60 & 0.07 & 0.06 & 0.02 & 0.14 & 0.01 \\
Whole & \textbf{69.76} & 7.11 & 2.65 & 0.02 & 1.92  & 4.68 & 0.00 \\ \hline
\end{tabular}}
\end{table}
\vspace{-0.75em}
\subsection{Experimental Setup}
\label{sec:setup}
This subsection provides comprehensive overviews of our experimental setup, including the datasets, dataset distillation methods, adversarial attack methods, network architectures, and implementation details.

\subsubsection{Datasets}
\label{sec:Datasets}

Our experiments utilize a range of datasets: CIFAR-10, CIFAR-100 \cite{krizhevsky2009learning}, TinyImageNet \cite{le2015tiny}, and ImageNet-1K \cite{russakovsky2015imagenet}. 
CIFAR-10 and CIFAR-100 each consist of $50,000$ training images and $10,000$ test images, with a resolution of $32 \times 32$ pixels. 
TinyImageNet, a downscaled version of ImageNet, comprises $100,000$  training images, $10,000$ validation images, and $10,000$  test images, each $64 \times 64$ in size. 
The original ImageNet-1K consists of $1,000$ categories of much higher resolution images with over $1,000,000$ in total.
For subsets of ImageNet-1K, following the experimental settings from MTT \cite{Cazenavette:MTT}, we employ subsets like ImageNette, ImageWoof, ImageFruit, ImageMeow, ImageSquawk, and ImageYellow, each with $10$ classes and resized to $128 \times 128$.

\subsubsection{Distillation Methods}
\label{sec:distillation}
In our study, we evaluate representative distillation methods, selected for their prominence and varied approaches in recent research. 
These include DC \cite{Zhao:DC}, DSA \cite{Zhao:DSA}, DM \cite{Zhao:DM}, MTT \cite{Cazenavette:MTT}, TESLA \cite{cui2023scaling}, IDM \cite{zhao2023improved}, DREAM \cite{Liu:DREAM}, SRe2L \cite{yin2023squeeze} and D4M \cite{su2024d4m}. 
These methods span several optimization approaches: gradient matching (DC, DSA), distribution matching (DM, IDM), trajectory matching (MTT, TESLA), and optimization-based methods (DREAM). 
Furthermore, we explore innovative techniques such as the decoupling method in SRe2L and the generative model-based approach in D4M, enhancing our understanding of their effectiveness in real-world applications.

\subsubsection{Adversarial Attack}
\label{sec:attacks}
By injecting noise into the test set, we evaluate the adversarial robustness of the models trained on distilled data against perturbations.
We apply $6$ widely used attacks in a non-targeted mode: FGSM\cite{goodfellow2014explaining}, PGD \cite{madry2017towards}, CW \cite{carlini2017towards},VMIFGSM\cite{wang2021vmifgsm}, Jitter\cite{schwinn2023jitter} and AutoAttack\cite{liu2022practical:autoattack}. This collection includes the most widely used benchmarking attacks (FGSM, PGD, CW and AutoAttack) and their improved version (VMIFGSM and Jitter).

\begin{table*}[p]
  \caption{Robust accuracy after AutoAttack attack. In the second column, the value represents the magnitude of perturbation (divided by $255$).}
\label{tab:auto}
\setlength{\tabcolsep}{2.5pt} 
\renewcommand{\arraystretch}{1.1} 
\centering
\tiny
\resizebox{\textwidth}{!}{
\begin{tabular}{cc|ccccc|ccccc|ccccc}
\toprule
\multicolumn{2}{c|}{Dataset} &\multicolumn{5}{c|}{\textbf{CIFAR-10}} & \multicolumn{5}{c|}{\textbf{CIFAR-100}} & \multicolumn{5}{c}{\textbf{TinyImageNet}} \\
\hline
\multicolumn{2}{c|}{IPC}   & 1 & 5 & 10 & 30 & 50 & 1 & 5 & 10 & 30 & 50 & 1 & 5 & 10 & 30 & 50 \\ \specialrule{0.8pt}{0pt}{0pt}
\rowcolor{nbarrier}\multirow{5}{*}{\rotatebox{90}{\textcolor{nbarrier}{$\blacksquare$} DC}} & \textbf{Clean} & \textbf{29.73} & \textbf{41.23} & \textbf{46.07} & \textbf{53.65} & \textbf{55.08} & \textbf{12.82} & \textbf{21.97} & \textbf{25.66} & \textbf{30.82} & \textbf{30.19} & \textbf{5.49} & \textbf{9.70} & \textbf{11.74} & \textbf{13.51} & \textbf{11.97} \\
& 1 & 23.48 & 28.66 & 28.67 & 31.98 & 30.03 & 7.98 & 13.49 & 15.50 & 14.18 & 10.94 & 1.91 & 3.54 & 3.44 & 1.87 & 1.62  \\
 & 2 & 18.19 & 18.34 & 16.00 & 15.78 & 13.38 & 4.68 & 8.23 & 9.69 & 5.92 & 3.92 & 0.52 & 1.27 & 0.81 & 0.16 & 0.27 \\
 & 3 & 13.93 & 10.70 & 8.07 & 6.53 & 5.05 & 2.68 & 5.46 & 5.75 & 2.60 & 1.52 & 0.11 & 0.37 & 0.18 & 0.04 & 0.04  \\ 
  & 4 & 10.25 & 6.09 & 3.67 & 2.26 & 1.49 & 1.58 & 3.54 & 3.45 & 1.06 & 0.54 & 0.03 & 0.09 & 0.05 & 0.00 & 0.00  \\
 \hline
\rowcolor{nbicycle}\multirow{5}{*}{\rotatebox{90}{\textcolor{nbicycle}{$\blacksquare$} DSA}} & \textbf{Clean} & \textbf{29.27} & \textbf{48.20} & \textbf{52.93} & \textbf{56.41} & \textbf{61.14} & \textbf{14.38} & \textbf{27.13} & \textbf{32.94} & \textbf{37.29} & \textbf{42.87} & \textbf{5.47} & \textbf{13.99} & \textbf{17.47} & \textbf{17.12} & \textbf{21.89} \\
& 1 & 22.35 & 31.51 & 33.70 & 29.54 & 31.85 & 8.99 & 15.83 & 17.50 & 13.65 & 18.20 & 2.21 & 5.19 & 5.89 & 4.31 & 4.34  \\
 & 2 & 16.47 & 19.22 & 19.11 & 12.89 & 12.95 & 5.92 & 8.98 & 8.82 & 4.87 & 7.63 & 0.92 & 1.83 & 2.14 & 0.91 & 0.83  \\
 & 3 & 11.42 & 10.63 & 9.54 & 4.48 & 4.36 & 3.76 & 5.27 & 4.63 & 1.92 & 3.14 & 0.42 & 0.60 & 0.62 & 0.29 & 0.14 \\ 
  & 4 & 7.98 & 5.33 & 4.25 & 1.42 & 1.30 & 2.39 & 3.06 & 2.40 & 0.73 & 1.46 & 0.12 & 0.14 & 0.18 & 0.02 & 0.02 \\
 \hline
\rowcolor{nbus}\multirow{5}{*}{\rotatebox{90}{\textcolor{nbus}{$\blacksquare$} DM}} & \textbf{Clean} & \textbf{26.75} & \textbf{42.78} & \textbf{49.81} & \textbf{59.97} & \textbf{63.12} & \textbf{11.71} & \textbf{23.90} & \textbf{29.92} & \textbf{38.48} & \textbf{43.56} & \textbf{4.06} & \textbf{9.76} & \textbf{14.16} & \textbf{21.03} & \textbf{21.29} \\
& 1 & 21.78 & 25.58 & 28.71 & 36.13 & 38.07 & 7.69 & 11.85 & 14.35 & 15.54 & 15.32 & 1.95 & 2.45 & 3.84 & 4.29 & 3.45   \\
 & 2 & 17.91 & 14.10 & 14.18 & 19.24 & 18.23 & 5.18 & 5.61 & 6.20 & 5.55 & 5.15 & 0.88 & 0.77 & 0.99 & 0.87 & 0.48 \\
 & 3 & 14.16 & 7.20 & 5.83 & 9.47 & 7.90 & 3.41 & 2.70 & 2.71 & 2.01 & 1.79 & 0.40 & 0.18 & 0.26 & 0.15 & 0.03  \\ 
  & 4 & 11.17 & 3.56 & 2.40 & 4.02 & 2.89 & 2.22 & 1.32 & 1.25 & 0.78 & 0.64 & 0.21 & 0.03 & 0.06 & 0.02 & 0.01  \\
 \hline
\rowcolor{ncar}\multirow{5}{*}{\rotatebox{90}{\textcolor{ncar}{$\blacksquare$} MTT}} & \textbf{Clean} & \textbf{45.74} & \textbf{57.19} & \textbf{60.98} & \textbf{66.23} & \textbf{70.45} & \textbf{20.30} & \textbf{34.81} & \textbf{37.84} & \textbf{42.65} & \textbf{44.34} & \textbf{8.91} & \textbf{14.86} & \textbf{19.93} & \textbf{23.05} & \textbf{26.16} \\
& 1 & 29.25 & 33.92 & 35.79 & 38.55 & 40.12 & 10.96 & 16.03 & 16.48 & 16.26 & 17.45 & 2.78 & 3.02 & 3.67 & 3.12 & 2.61   \\
 & 2 & 16.41 & 17.39 & 17.95 & 17.14 & 17.41 & 5.79 & 6.97 & 6.40 & 5.04 & 5.65 & 0.76 & 0.50 & 0.57 & 0.32 & 0.26 \\
 & 3 & 8.32 & 7.50 & 7.45 & 6.03 & 5.57 & 3.10 & 3.22 & 2.31 & 1.60 & 1.87 & 0.15 & 0.04 & 0.11 & 0.04 & 0.02 \\ 
  & 4 & 3.88 & 3.07 & 2.47 & 1.82 & 1.31 & 1.67 & 1.42 & 0.83 & 0.55 & 0.66 & 0.06 & 0.02 & 0.01 & 0.00 & 0.01 \\
 \hline
\rowcolor{npedestrian}\multirow{5}{*}{\rotatebox{90}{\textcolor{npedestrian}{$\blacksquare$} TESLA}} & \textbf{Clean} & \textbf{47.07} & \textbf{57.23} & \textbf{61.78} & \textbf{67.08} & \textbf{68.83} & \textbf{20.00} & \textbf{31.65} & \textbf{34.99} & \textbf{39.16} & \textbf{45.92} & \textbf{7.40} & \textbf{16.01} & \textbf{19.45} & \textbf{23.71} & \textbf{26.63} \\
& 1 & 30.05 & 34.57 & 35.00 & 38.92 & 35.73 & 10.79 & 15.39 & 10.84 & 13.34 & 17.76 & 1.92 & 2.71 & 2.83 & 3.25 & 2.31  \\
 & 2 & 17.46 & 17.09 & 15.12 & 17.75 & 14.09 & 5.45 & 6.98 & 3.00 & 3.85 & 5.83 & 0.48 & 0.35 & 0.31 & 0.46 & 0.27 \\
 & 3 & 9.61 & 7.05 & 5.67 & 6.43 & 4.11 & 2.92 & 3.27 & 0.82 & 1.01 & 1.81 & 0.11 & 0.03 & 0.01 & 0.07 & 0.01 \\ 
  & 4 & 5.00 & 2.39 & 1.51 & 1.79 & 1.12 & 1.63 & 1.43 & 0.23 & 0.39 & 0.63 & 0.00 & 0.00 & 0.00 & 0.01 & 0.00 \\
 \hline
\rowcolor{nterrain}\multirow{5}{*}{\rotatebox{90}{\textcolor{nterrain}{$\blacksquare$} SRe2L}} & \textbf{Clean} & \textbf{13.49} & \textbf{31.06} & \textbf{37.53} & \textbf{54.88} & \textbf{63.28} & \textbf{4.68} & \textbf{28.92} & \textbf{39.64} & \textbf{51.33} & \textbf{53.95} & \textbf{6.28} & \textbf{18.38} & \textbf{26.92} & \textbf{39.49} & \textbf{43.24} \\
& 1 & 8.82 & 21.03 & 22.91 & 30.99 & 35.34 & 2.44 & 9.29 & 12.37 & 14.30 & 15.28 & 2.45 & 3.68 & 5.52 & 8.84 & 10.89   \\
 & 2 & 5.66 & 13.46 & 12.56 & 14.31 & 14.39 & 1.39 & 2.29 & 2.49 & 2.37 & 2.06 & 1.03 & 0.64 & 0.98 & 1.53 & 1.73 \\
 & 3 & 4.20 & 7.68 & 6.10 & 5.25 & 4.51 & 0.96 & 0.76 & 0.62 & 0.40 & 0.33 & 0.48 & 0.15 & 0.22 & 0.27 & 0.38 \\ 
  & 4 & 3.47 & 3.92 & 2.75 & 1.57 & 1.05 & 0.73 & 0.20 & 0.14 & 0.09 & 0.06 & 0.28 & 0.02 & 0.08 & 0.05 & 0.06 \\
 \hline
 \rowcolor{grey}\multirow{5}{*}{\rotatebox{90}{\textcolor{grey}{$\blacksquare$} IDM}} & \textbf{Clean} & \textbf{46.14} & \textbf{54.45} & \textbf{58.84} & \textbf{65.83} & \textbf{67.82} & \textbf{23.90} & \textbf{40.02} & \textbf{45.33} & \textbf{45.83} & \textbf{46.13} & \textbf{10.42} & \textbf{22.39} & \textbf{22.4} & \textbf{24.96} & \textbf{26.89} \\
 & 1 & 32.28 & 39.67 & 38.61 & 43.25 & 42.21 & 12.56 & 19.07 & 20.54 & 17.44 & 15.38 & 4.47 & 6.40 & 4.59 & 3.88 & 5.78 \\
 & 2 & 20.82 & 25.85 & 22.63 & 23.43 & 21.92 & 6.43 & 8.64 & 8.33 & 5.74 & 4.31 & 1.80 & 1.94 & 1.04 & 0.77 & 1.35 \\
 & 3 & 12.46 & 15.90 & 11.61 & 11.36 & 9.68 & 3.51 & 4.03 & 3.24 & 2.03 & 1.40 & 0.74 & 0.47 & 0.23 & 0.05 & 0.28 \\
  & 4 & 6.98 & 8.65 & 5.38 & 4.61 & 3.57 & 1.85 & 1.86 & 1.28 & 0.75 & 0.42 & 0.34 & 0.13 & 0.05 & 0.00 & 0.03 \\

 \hline
 \rowcolor{purple}\multirow{5}{*}{\rotatebox{90}{\textcolor{purple}{$\blacksquare$} DREAM}} & \textbf{Clean} & \textbf{44.66}&\textbf{57.07}&\textbf{63.25}&\textbf{67.75}&\textbf{68.96}&\textbf{24.95} &\textbf{36.17} &\textbf{42.36} &\textbf{44.46} &\textbf{46.84} &\textbf{7.59} &\textbf{17.58} &\textbf{18.88} &\textbf{19.91} &\textbf{17.46} \\
 & 1 & 32.08 & 36.20 & 38.44 & 37.12 & 40.54 & 13.50 & 14.89 & 17.22 & 13.35 & 13.84 & 2.28 & 4.27 & 3.16 & 2.22 & 3.63 \\
 & 2 & 21.40 & 20.41 & 20.23 & 17.16 & 18.81 & 7.58 & 5.69 & 6.48 & 3.96 & 4.00 & 0.52 & 0.99 & 0.58 & 0.19 & 0.61 \\
 & 3 & 12.79 & 9.82 & 9.06 & 6.92 & 7.26 & 4.46 & 2.31 & 2.63 & 1.22 & 1.17 & 0.16 & 0.15 & 0.08 & 0.00 & 0.08 \\
  & 4 & 7.51 & 4.22 & 3.82 & 2.24 & 2.64 & 2.49 & 1.04 & 0.95 & 0.47 & 0.40 & 0.05 & 0.01 & 0.00 & 0.00 & 0.01 \\
 \hline

 \rowcolor{cyan}\multirow{5}{*}{\rotatebox{90}{\textcolor{cyan}{$\blacksquare$} D4M}} & \textbf{Clean} & \textbf{23.39}&\textbf{42.34}&\textbf{48.16}&\textbf{63.53}&\textbf{69.81}&\textbf{11.27} &\textbf{31.90} &\textbf{40.12} &\textbf{48.15} &\textbf{51.10} &\textbf{2.16} &\textbf{6.94} &\textbf{14.25} &\textbf{35.80} &\textbf{42.89} \\
 & 1 & 14.75 &26.11 &32.19 &39.75 &43.00 &5.60 &12.21 &13.76 &14.94 &13.49 &0.82 &1.72 &2.63 &6.70 &10.49 \\
 & 2 & 7.88 &14.10 &18.51 &20.96 &21.77 &2.54 &3.79 &3.70 &3.45 &2.56 &0.46 &0.77 &0.72 &1.31 &2.13 \\
 & 3 & 4.24 &6.75 &10.23 &9.48 &9.45 &1.22 &1.11 &0.91 &0.85 &0.48 &0.26 &0.35 &0.22 &0.34 &0.53 \\
  & 4 & 1.93 &3.06 &5.07 &3.58 &3.50 &0.63 &0.29 &0.31 &0.23 &0.12 &0.18 &0.17 &0.11 &0.09 &0.24 \\
 \hline
 
\multicolumn{2}{c|}{\multirow{2}{*}{\textbf{Whole}}} 
&\textbf{1}&\textbf{2}&\textbf{3}&\textbf{4}& \textbf{Clean} &\textbf{1}&\textbf{2}&\textbf{3}&\textbf{4}& \textbf{Clean} &\textbf{1}&\textbf{2}&\textbf{3}&\textbf{4}& \textbf{Clean} \\& 
&41.76&12.44&2.49&0.49&84.45&7.26&0.71&0.10&0.03&55.74&9.65&1.74&0.35&0.06&38.77 \\

 \bottomrule
\end{tabular}
}
\end{table*}

\begin{table*}[]
\caption{Robust accuracy of ImageNet-subsets after AutoAttack attack. The distilled data is acquired by MTT. In the first column, the value represents the magnitude of perturbation
(divided by 255).}
\label{tab:a-subauto}
\centering
\setlength{\tabcolsep}{2.5pt} 
\renewcommand{\arraystretch}{1.1}
\resizebox{\textwidth}{!}{%
\begin{tabular}{c|cccccccccccccccccc}
\toprule
      \multirow{2}{*}{IPC}               & \multicolumn{6}{c|}{\cellcolor{nbarrier}\textbf{ImageNette}}        & \multicolumn{6}{c|}{\cellcolor{nbicycle}\textbf{ImageWoof}}         & \multicolumn{6}{c}{\cellcolor{nbus}\textbf{ImageFruit}}        \\
 & 1 & 5 & 10 & 30 & 50 & \multicolumn{1}{c|}{Whole} & 1 & 5 & 10 & 30 & 50 & \multicolumn{1}{c|}{Whole} & 1 & 5 & 10 & 30 & 50 & Whole \\ \hline
\rowcolor{grey!40}\textbf{Clean} & \textbf{48.20} & \textbf{62.00} & \textbf{66.40} & \textbf{66.60} & \textbf{67.60} & \multicolumn{1}{c|}{\textbf{86.40}} & \textbf{30.40} & \textbf{35.20} & \textbf{38.00} & \textbf{38.80} & \textbf{39.40} & \multicolumn{1}{c|}{\textbf{67.20}} & \textbf{25.00} & \textbf{41.00} & \textbf{42.20} & \textbf{44.40} & \textbf{44.60} & \textbf{67.40} \\
1 & 29.40 & 40.00 & 40.40 & 38.00 & 39.40 & \multicolumn{1}{c|}{53.40} & 13.60 & 9.20 & 12.80 & 11.20 & 10.60 & \multicolumn{1}{c|}{12.40} & 11.40 & 16.00 & 20.00 & 18.80 & 18.00 & 23.80 \\
2 & 16.60 & 22.00 & 20.60 & 18.00 & 18.00 & \multicolumn{1}{c|}{17.60} & 3.60 & 2.80 & 3.20 & 1.80 & 1.60 & \multicolumn{1}{c|}{0.40} & 5.40 & 4.40 & 7.80 & 5.00 & 6.80 & 5.20 \\
3 & 8.60 & 9.20 & 10.00 & 7.80 & 8.20 & \multicolumn{1}{c|}{3.80} & 2.40 & 0.80 & 0.80 & 0.40 & 0.00 & \multicolumn{1}{c|}{0.00} & 3.00 & 2.00 & 1.20 & 2.20 & 2.00 & 1.00 \\
4 & 4.00 & 4.60 & 3.40 & 2.20 & 1.20 & \multicolumn{1}{c|}{0.40} & 1.20 & 0.40 & 0.20 & 0.20 & 0.00 & \multicolumn{1}{c|}{0.00} & 1.40 & 0.80 & 0.20 & 0.80 & 1.00 & 0.00 \\ \toprule
      \multirow{2}{*}{IPC}                 & \multicolumn{6}{c|}{\cellcolor{ncar}\textbf{ImageMeow}}         & \multicolumn{6}{c|}{\cellcolor{npedestrian}\textbf{ImageSquawk}}       & \multicolumn{6}{c}{\cellcolor{nterrain}\textbf{ImageYellow}}       \\ 
 & 1 & 5 & 10 & 30 & 50 & \multicolumn{1}{c|}{Whole} & 1 & 5 & 10 & 30 & 50 & \multicolumn{1}{c|}{Whole} & 1 & 5 & 10 & 30 & 50 & Whole \\ \hline
\rowcolor{grey!40}\textbf{Clean} & \textbf{31.00} & \textbf{41.40} & \textbf{44.40} & \textbf{44.20} & \textbf{44.20} & \multicolumn{1}{c|}{\textbf{69.00}} & \textbf{39.00} & \textbf{52.40} & \textbf{55.60} & \textbf{56.20} & \textbf{59.20} & \multicolumn{1}{c|}{\textbf{86.40}} & \textbf{44.60} & \textbf{59.20} & \textbf{63.40} & \textbf{65.40} & \textbf{66.20} & \textbf{84.40} \\
1 & 11.80 & 15.00 & 16.00 & 10.40 & 10.80 & \multicolumn{1}{c|}{14.00} & 20.40 & 27.80 & 26.00 & 25.20 & 24.60 & \multicolumn{1}{c|}{48.00} & 25.40 & 33.00 & 38.20 & 32.60 & 35.60 & 48.00 \\
2 & 4.40 & 4.20 & 4.80 & 3.80 & 2.60 & \multicolumn{1}{c|}{1.40} & 10.20 & 13.40 & 10.80 & 9.00 & 10.20 & \multicolumn{1}{c|}{17.80} & 13.60 & 19.60 & 16.60 & 17.00 & 16.40 & 17.40 \\
3 & 2.00 & 0.80 & 1.20 & 0.80 & 0.60 & \multicolumn{1}{c|}{0.40} & 4.40 & 6.40 & 4.20 & 4.20 & 4.20 & \multicolumn{1}{c|}{6.20} & 8.00 & 7.40 & 9.00 & 7.40 & 8.20 & 5.00 \\
4 & 0.00 & 0.20 & 0.60 & 0.60 & 0.60 & \multicolumn{1}{c|}{0.20} & 1.20 & 2.20 & 2.00 & 1.60 & 1.40 & \multicolumn{1}{c|}{2.80} & 2.20 & 4.00 & 4.40 & 2.80 & 3.00 & 1.60 \\
\bottomrule
\end{tabular}%
}
\end{table*}

\subsubsection{Network Architecture}
\label{sec:Network Architecture}

Following the prior works\cite{Zhao:DC,Zhao:DM,Cui:dcbench}, we utilize a default ConvNet comprises three identical convolutional blocks in our experiments. Each block contains a $128$-kernel convolution layer, instance normalization, and average pooling.
For TinyImageNet and ImageNet-subsets, we employ ConvNetD4 and ConvNetD5 to accommodate larger resolution, which replicate the convolutional block one and two additional times.
While this simple convolutional network may not be as susceptible to attack algorithms, our focus lies not in achieving maximum model robustness but rather in evaluating the robustness of the distilled data and the distillation method itself. 
Moreover, previous studies \cite{Cui:dcbench} suggest that maintaining consistent architecture throughout the distillation and training processes is optimal. 
Thus, we ensure that both the synthesis of synthetic data and subsequent operations maintain architecture consistency with distillation.
We apply ResNet18 with batch normalization as the base model for SRe2L in all stages. The $7\times7$ Conv layer is replaced by $3\times3$ Conv layer and the max-pooling layer is discarded for TinyImageNet and smaller datasets following the settings in SRe2L \cite{yin2023squeeze}. We utilize the same ConvNet and modified ResNet18 architecture in distilling CIFAR and TinyImageNet datasets with D4M accordingly for a fair comparison.

To make our evaluation more extensive, we trained 3 more different network architectures on distilled datasets,including  ResNet18 \cite{resnet18}, VGG \cite{vgg} and MobileNet \cite{mobile}, and evaluated their clean and robust accuaries. We show the results in supplementary materials.

\vspace{-0.75em}
\subsection{Implement Details}
\label{sec:implement}
This section provides detailed descriptions of the parameters validated in the three distinct stages of our benchmarking process.

\subsubsection{Distillation Stage} 
In the distillation stage, we adhere to the parameters utilized in prior distillation methods~\footnote{We set multi-formation factor to 2 for both IDM and DREAM.}. 
Contrary to the ZCA whitening technique applied in the methodologies of MTT \cite{Cazenavette:MTT} and TESLA \cite{cui2023scaling}, which is intended to accelerate convergence, we opted to exclude this step. 
We instead modified other parameters, such as learning rate and synthetic steps, to better suit our experimental framework. 
These adjustments are detailed in the supplementary materials.
Regarding compression rates, we maintain consistency with previous studies\cite{Zhao:DSA,Zhao:DM,Cui:dcbench} by adopting IPC values of 1, 10, and 50. 
Additionally, we extend our experiments to include IPC values of 5 and 30. 
This expansion allows for a more comprehensive analysis of data variation trends across different compression rates.

\subsubsection{Training Stage} 
After obtaining the distilled data, we train five models on each distilled dataset using an SGD optimizer from scratch, with each model undergoing $1,000$ epochs of training. 
Training parameters are standardized across all datasets, with the exception of SRe2L and D4M, which follow their specific FKD training strategy.
Methods utilizing DSA augmentation maintain consistency with their respective methods during training.
Subsequently, we select models that achieve the highest validation accuracy on the original test set for subsequent adversarial attacks. 

\subsubsection{Attacking Stage} 
In this stage, we maintain uniformity across all trained models by setting identical parameters for a fair comparison. 
FGSM, being a one-step attack, does not require adjustments to step size or iteration number. 
We set the perturbation budget $|\varepsilon| = \frac{2}{255}, \frac{4}{255}, \frac{8}{255}$ for all methods except AutoAttack.
AutoAttack, being parameter-free, does not require specification of step size or iteration number, providing final attack results and individual effects of each sub-algorithm using their default modes. 
Given the robust capabilities of AutoAttack, we test scenarios with perturbation bounds set to $|\varepsilon| = \frac{1}{255}, \frac{2}{255}, \frac{3}{255}, \frac{4}{255}$.
We classify samples as negative examples if the model's output contradicts the original label of the test sample, without considering the specific class to which it has been incorrectly assigned.

\begin{table*}[h]
\centering
\caption{Impact of three different components including data augmentation, downsampling factor and clustering number. Perturbation of adversarial attacks is set to $|\varepsilon|=2/255$.}
\label{tab:component}

  \begin{minipage}[c]{0.45\linewidth}

  \centering
\begin{subtable}[]{\textwidth}

\centering

\renewcommand\arraystretch{1.4}
\caption{Clean and robust accuracy under different data augmentations using DSA.}
\vspace{1em}
\label{tab:aug}
\resizebox{0.9\textwidth}{!}{
\begin{tabular}{ccccc}
\hline
\textbf{Aug} & \textbf{Clean} & \textbf{PGD} & \textbf{CW} & \textbf{Auto} \\ \hline
Color & 48.14 & 19.06 & 18.28 & 17.87 \\
Crop & 48.48 & 17.26 & 17.20 & 16.73 \\
Cutout & 48.02 & 17.57 & 16.89 & 16.41 \\
Flip & 49.16 & 22.73 & 22.00 & 21.76 \\
Rotate & 46.66 & 19.42 & 18.83 & 18.45 \\
Scale & 47.04 & 19.06 & 18.5 & 18.14 \\
DSA & 52.93 & 20.01 & 19.83 & 19.11 \\
None(DC) & 46.07 & 18.27 & 16.50 & 16.00 \\ \hline
\end{tabular}}
\end{subtable}
\end{minipage}
\qquad
\begin{minipage}[c]{0.45\linewidth}

\centering
\begin{subtable}[]{\textwidth}
\centering
\renewcommand\arraystretch{1.1}
\caption{Clean and robust accuracy under different downsampling factor $f$ using IDC.}
\label{tab:downsampling}
\vspace{-0.5em}
  \resizebox{0.77\textwidth}{!}{
\begin{tabular}{ccccc}
\hline
\textbf{$f$} & \textbf{Clean} & \textbf{PGD} & \textbf{CW} & \textbf{Auto} \\ \hline
1 & 51.81 & 21.88 & 21.66 & 21.27 \\
2 & 62.06 & 21.24 & 21.24 & 20.53 \\
3 & 63.54 & 15.56 & 15.69 & 14.88 \\
4 & 61.21 & 9.59 & 9.73 & 8.91 \\\hline
\end{tabular}}%
\end{subtable}

\begin{subtable}[]{\textwidth}
\centering
\renewcommand\arraystretch{1.1}
\vspace{1em}
\caption{Clean and robust accuracy under different clustering number $N$ using DREAM.}
\label{tab:cluster}
\vspace{-0.5em}
\resizebox{0.77\textwidth}{!}{
\begin{tabular}{ccccc}
\hline
\textbf{$N$} & \textbf{Clean} & \textbf{PGD} & \textbf{CW} & \textbf{Auto} \\ \hline
32 & 58.61 & 21.06 & 21.08 & 20.56 \\
64 & 61.46 & 22.58 & 22.48 & 21.88 \\
128 & 63.28 & 21.88 & 21.89 & 21.35 \\
256 & 62.99 & 21.60 & 21.55 & 21.08\\ \hline
\end{tabular}}%
\end{subtable}
\end{minipage}
\end{table*}

\vspace{-0.75em}
\subsection{Experimental Results }
\label{sec:results}

In Table \ref{table:benchmark2/255} and Table \ref{table:benchmark4/255}-\ref{table:benchmark8/255}, we present adversarial robustness and original test accuracies for CIFAR-10/100 and TinyImageNet. Results for ImageNet subsets and ImageNet-1K are displayed in Table  \ref{table:imagenetsubset2/255}-\ref{table:imagenet1k} and Table  \ref{table:imagenetsubset4/255}-\ref{table:imagenetsubset8/255}. The results of AutoAttack are listed separately in Table \ref{tab:auto} and \ref{tab:a-subauto}. 
Regarding the attacking iteration number, we report the results under the setting of $10$ iterations in the tables. 
Additionally, evaluations were conducted with the attacking iteration number set to $5$, but no significant difference compared to $10$ iterations was observed. Perturbation step size is fixed at $|\varepsilon|=\frac{2}{255}$

To provide a more intuitive visualization of these results, we have plotted the accuracy drop rate ($DR$) trends based on Table \ref{table:benchmark2/255} in Figure \ref{fig:droprate}.
$DR$ is formulated as follows:
\begin{equation}\label{eq:droprate}
DR=\frac{Acc_{original}-Acc_{robust}}{Acc_{original}} \times 100\%
\end{equation}

It stands for the ratio of the decrease in model evaluation accuracy after perturbing the test set to the original test accuracy. 
Such a ratio form provides a relatively fair comparison of the robustness of models across different settings, as it somehow mitigates the differences in original test accuracy (especially those caused by IPC). 
A higher value of this ratio indicates that the model is more sensitive to adversarial attacks, resulting in lower robustness of the distilled data. Conversely, a lower $DR$ suggests that models trained using such distilled data are more robust to adversarial attacks.

We have adopted this normalized approach because it enables a more visually intuitive comparison of robustness across different conditions of clean accuracy.
By leveraging $DR$, we discovered that distilled data exhibits similar trends in robustness variations under different attacks. Taking PGD as an example for analysis, we uncovered some intriguing patterns.

Our analysis has revealed distinct patterns across various distillation methods. 
For instance, DC consistently demonstrates a lower drop rate compared to nearly all other methods. 
Being a gradient-matching-based method as well, DSA efficiently improves clean accuracy while weakens robustness to some degree. We will discuss this in the following section. 
Notably, DM consistently achieves the lowest drop rate when IPC is set to 1. 
As for methods based on trajectory matching (such as MTT and TESLA), their drop rates exhibit a smooth and closely resembling trend to that of the original dataset. 
For ImageNet-1K, despite SRe2L achieving remarkable performance in terms of accuracy, we discovered that models trained using it exhibit relatively low robustness. In comparison, MTT demonstrates better robustness performance on its subsets.
DREAM and IDM seem to show a similar trend, which may attribute to the multi-formation setting based on \cite{kim2022idc}.

Our primary validated observation reveals that the accuracy $DR$ of these methods shows an upward trend across the three datasets as IPC increases. 
This trend closely mirrors the variations observed between accuracy and IPC. 
Such phenomenon becomes even more evident when perturbation grows to be very substantial. For example, Table \ref{table:imagenetsubset8/255} reveals that $IPC=1$ achieves significant highest robust accuracy than other IPCs in most cases. 
The inverse relationship between robustness and IPC suggests that merely increasing the number of distilled images may not be wise. 
Opting for a lower IPC could potentially yield more robust models. 
For instance, while the original accuracy notably improves with increasing IPC in most cases, Table \ref{table:benchmark2/255} demonstrates that the distilled TinyImageNet obtained using MTT experiences a consistent decline in robust accuracy following the PGD attack. 
This contradictory trend strongly indicates a decline in robustness, particularly evident under high IPC settings ($30$ and $50$). 
This finding is intriguing, as insufficient data often leads to overfitting.
Notably, DC-Bench \cite{Cui:dcbench} illustrated that simply elevating IPC does not necessarily enhance model accuracy under high IPC settings. 
Our benchmark offers a complementary viewpoint that robustness also plays an important role in dataset distillation and is correlated with the synthetic scale.

Another noteworthy observation coming from our benchmark is that models trained using distilled CIFAR-10, CIFAR-100, and TinyImageNet datasets (particularly when IPC is relatively low) demonstrate superior robustness compared to those trained on the original dataset. 
We posit that this phenomenon may be attributed to the distillation process introducing refined knowledge.

\begin{table*}[p]
  \caption{Clean and robust accuracies of models trained on distilled CIFAR-10, CIFAR-100 and TinyImageNet. The perturbation budget is set to $|\varepsilon|=4/255$.}
\label{table:benchmark4/255}
\setlength{\tabcolsep}{2.5pt} 
\tiny
\renewcommand{\arraystretch}{1} 
\centering
\resizebox{\textwidth}{!}{
\begin{tabular}{cc|ccccc|ccccc|ccccc}
\toprule
\multicolumn{2}{c|}{Dataset} &\multicolumn{5}{c|}{\textbf{CIFAR-10}} & \multicolumn{5}{c|}{\textbf{CIFAR-100}} & \multicolumn{5}{c}{\textbf{TinyImageNet}} \\
\hline
\multicolumn{2}{c|}{IPC}   & 1 & 5 & 10 & 30 & 50 & 1 & 5 & 10 & 30 & 50 & 1 & 5 & 10 & 30 & 50 \\ \specialrule{0.8pt}{0pt}{0pt}
\rowcolor{nbarrier}\multirow{6}{*}{\rotatebox{90}{\textcolor{nbarrier}{$\blacksquare$} DC}} & \textbf{Clean} & \textbf{29.73} & \textbf{41.23} & \textbf{46.07} & \textbf{53.65} & \textbf{55.08} & \textbf{12.82} & \textbf{21.97} & \textbf{25.66} & \textbf{30.82} & \textbf{30.19} & \textbf{5.49} & \textbf{9.70} & \textbf{11.74} & \textbf{13.51} & \textbf{11.97} \\
 & FGSM & 13.32 &9.42 &7.32 &5.81 &4.69 &3.40 &5.65 &5.80 &2.98 &1.92 &0.30 &0.63 &0.32 &0.06 &0.07 \\
 & PGD & 12.15 &7.15 &4.94 &2.82 &1.85 &2.82 &4.42 &4.20 &1.44 &0.76 &0.17 &0.33 &0.14 &0.01 &0.00 \\
 & CW & 10.58 &6.61 &4.25 &2.68 &1.77 &1.72 &3.78 &4.00 &1.20 &0.65 &0.05 &0.09 &0.08 &0.01 &0.00 \\
 & VMI &12.18 &7.36 &5.14 &3.03 &1.95 &2.81 &4.51 &4.34 &1.58 &0.84 &0.17 &0.35 &0.13 &0.02 &0.00 \\
 & Jitter & 11.29 &9.69 &8.18 &7.55 &6.51 &1.81 &3.89 &4.01 &1.51 &1.04 &0.05 &0.18 &0.11 &0.03 &0.03 \\\hline
\rowcolor{nbicycle}\multirow{6}{*}{\rotatebox{90}{\textcolor{nbicycle}{$\blacksquare$} DSA}} & \textbf{Clean} & \textbf{29.27} & \textbf{48.20} & \textbf{52.93} & \textbf{56.41} & \textbf{61.14} & \textbf{14.38} & \textbf{27.13} & \textbf{32.94} & \textbf{37.29} & \textbf{42.87} & \textbf{5.47} & \textbf{13.99} & \textbf{17.47} & \textbf{17.12} & \textbf{21.89} \\
 & FGSM & 9.35 &8.70 &7.93 &3.65 &3.91 &3.81 &5.35 &5.01 &2.45 &3.81 &0.51 &0.82 &0.69 &0.40 &0.27 \\
 & PGD & 8.38 &5.94 &4.91 &1.69 &1.60 &3.25 &3.61 &2.83 &0.96 &1.77 &0.30 &0.29 &0.25 &0.10 &0.04 \\
 & CW & 8.52 &5.98 &4.96 &1.69 &1.59 &2.63 &3.46 &2.85 &0.90 &1.88 &0.17 &0.22 &0.24 &0.06 &0.03 \\
 & VMI & 8.54 &6.24 &5.11 &1.85 &1.69 &3.27 &3.83 &2.96 &0.97 &1.79 &0.33 &0.35 &0.31 &0.11 &0.03 \\
 & Jitter & 9.49 &9.90 &9.67 &5.55 &6.52 &2.57 &3.43 &3.02 &1.46 &2.20 &0.13 &0.23 &0.26 &0.10 &0.08 \\\hline
\rowcolor{nbus}\multirow{6}{*}{\rotatebox{90}{\textcolor{nbus}{$\blacksquare$} DM}} & \textbf{Clean} & \textbf{26.75} & \textbf{42.78} & \textbf{49.81} & \textbf{59.97} & \textbf{63.12} & \textbf{11.71} & \textbf{23.90} & \textbf{29.92} & \textbf{38.48} & \textbf{43.56} & \textbf{4.06} & \textbf{9.76} & \textbf{14.16} & \textbf{21.03} & \textbf{21.29} \\
 & FGSM & 12.10 &5.82 &4.84 &7.94 &7.13 &3.51 &2.75 &3.05 &2.79 &2.52 &0.46 &0.25 &0.37 &0.21 &0.13 \\
 & PGD & 11.41 &3.83 &2.78 &4.55 &3.39 &2.84 &1.62 &1.55 &1.00 &0.88 &0.33 &0.07 &0.07 &0.03 &0.02 \\
 & CW & 11.41 &4.14 &3.18 &4.79 &3.40 &2.38 &1.61 &1.63 &1.03 &0.87 &0.27 &0.10 &0.06 &0.03 &0.01 \\
 & VMI & 11.47 &4.01 &2.94 &4.69 &3.49 &2.91 &1.71 &1.62 &0.95 &0.80 &0.34 &0.08 &0.09 &0.04 &0.01 \\
 & Jitter & 11.55 &7.17 &7.01 &9.63 &9.08 &2.37 &1.61 &1.70 &1.26 &1.26 &0.29 &0.06 &0.07 &0.04 &0.03 \\\hline
\rowcolor{ncar}\multirow{6}{*}{\rotatebox{90}{\textcolor{ncar}{$\blacksquare$} MTT}} & \textbf{Clean} & \textbf{45.74} & \textbf{57.19} & \textbf{60.98} & \textbf{66.23} & \textbf{70.45} & \textbf{20.30} & \textbf{34.81} & \textbf{37.84} & \textbf{42.65} & \textbf{44.34} & \textbf{8.91} & \textbf{14.86} & \textbf{19.93} & \textbf{23.05} & \textbf{26.16} \\
 & FGSM & 8.59 &6.49 &7.78 &5.32 &4.89 &3.89 &4.04 &3.21 &2.65 &3.00 &0.35 &0.14 &0.22 &0.09 &0.06 \\
 & PGD & 6.23 &3.93 &4.54 &2.40 &1.84 &3.37 &2.19 &1.26 &0.77 &0.97 &0.15 &0.03 &0.03 &0.02 &0.02 \\
 & CW & 4.32 &3.39 &2.99 &2.29 &1.74 &1.85 &1.71 &1.11 &0.76 &0.97 &0.07 &0.02 &0.01 &0.00 &0.02 \\
 & VMI & 6.33 &4.13 &4.76 &2.57 &1.98 &3.37 &2.32 &1.42 &0.88 &1.02 &0.15 &0.04 &0.06 &0.02 &0.02 \\
 & Jitter & 8.55 &8.83 &9.12 &8.80 &8.36 &1.92 &2.08 &1.54 &1.20 &1.23 &0.07 &0.02 &0.03 &0.02 &0.03 \\\hline
\rowcolor{npedestrian}\multirow{6}{*}{\rotatebox{90}{\textcolor{npedestrian}{$\blacksquare$} TESLA}} & \textbf{Clean} & \textbf{47.07} & \textbf{57.23} & \textbf{61.78} & \textbf{67.08} & \textbf{68.83} & \textbf{20.00} & \textbf{31.65} & \textbf{34.99} & \textbf{39.16} & \textbf{45.92} & \textbf{7.40} & \textbf{16.01} & \textbf{19.45} & \textbf{23.71} & \textbf{26.63} \\
 & FGSM & 8.37 &5.83 &4.82 &5.85 &3.51 &3.28 &3.98 &1.63 &1.99 &2.74 &0.19 &0.07 &0.10 &0.14 &0.09 \\
 & PGD & 6.06 &3.38 &2.04 &2.44 &1.40 &2.52 &2.07 &0.44 &0.52 &0.83 &0.10 &0.01 &0.01 &0.03 &0.01 \\
 & CW & 5.30 &2.83 &1.89 &2.30 &1.40 &1.80 &1.73 &0.36 &0.55 &0.81 &0.00 &0.00 &0.00 &0.01 &0.00 \\
 & VMI & 6.31 &3.65 &2.31 &2.66 &1.55 &2.58 &2.15 &0.47 &0.62 &0.94 &0.10 &0.01 &0.01 &0.03 &0.01 \\
 & Jitter & 8.88 &8.44 &7.66 &9.06 &7.51 &1.88 &2.09 &0.60 &0.82 &1.29 &0.03 &0.00 &0.03 &0.02 &0.02 \\\hline
\rowcolor{nterrain}\multirow{6}{*}{\rotatebox{90}{\textcolor{nterrain}{$\blacksquare$} SRe2L}} & \textbf{Clean} & \textbf{13.49} & \textbf{31.06} & \textbf{37.53} & \textbf{54.88} & \textbf{63.28} & \textbf{4.68} & \textbf{28.92} & \textbf{39.64} & \textbf{51.33} & \textbf{53.95} & \textbf{6.28} & \textbf{18.38} & \textbf{26.92} & \textbf{39.49} & \textbf{43.24} \\
 & FGSM & 4.89 &6.71 &5.40 &6.41 &7.40 &1.45 &4.01 &5.35 &6.77 &7.49 &0.56 &0.37 &0.59 &1.14 &1.22 \\
 & PGD & 4.06 &4.42 &3.13 &2.22 &1.57 &1.20 &1.00 &0.63 &0.75 &0.62 &0.41 &0.06 &0.11 &0.18 &0.15 \\
 & CW & 3.65 &4.41 &3.23 &2.44 &2.16 &0.89 &0.57 &0.46 &0.40 &0.45 &0.40 &0.08 &0.18 &0.23 &0.21 \\
 & VMI & 3.99 &4.71 &3.42 &2.40 &1.80 &1.26 &0.87 &0.67 &0.56 &0.46 &0.42 &0.09 &0.13 &0.18 &0.21 \\
 & Jitter & 3.92 &8.74 &8.32 &9.92 &10.85 &0.84 &1.01 &1.81 &2.60 &2.54 &0.36 &0.06 &0.15 &0.21&0.35 \\\hline
 \rowcolor{grey}\multirow{6}{*}{\rotatebox{90}{\textcolor{grey}{$\blacksquare$} IDM}} & \textbf{Clean} & \textbf{46.14} & \textbf{54.45} & \textbf{58.84} & \textbf{65.83} & \textbf{67.82} & \textbf{23.90} & \textbf{40.02} & \textbf{45.33} & \textbf{45.83} & \textbf{46.13} & \textbf{10.42} & \textbf{22.39} & \textbf{22.4} & \textbf{24.96} & \textbf{26.89} \\
 & FGSM & 10.13 &12.78 &9.18 &8.63 &7.62 &3.54 &4.40 &3.78 &2.84 &2.13 &0.69 &0.57 &0.35 &0.19 &0.40 \\
 & PGD & 7.56 &9.46 &5.94 &5.63 &4.69 &2.32 &2.16 &1.69 &1.27 &0.77 &0.37 &0.28 &0.14 &0.04 &0.03 \\
 & CW & 7.60 &9.53 &6.02 &5.30 &4.09 &2.25 &2.20 &1.68 &1.06 &0.60 &0.39 &0.17 &0.09 &0.00 &0.04 \\
 & VMI & 7.96 &9.86 &6.19 &5.32 &4.04 &2.40 &2.24 &1.60 &0.93 &0.57 &0.43 &0.21 &0.07 &0.00 &0.04 \\
 & Jitter & 11.45 &14.24 &11.65 &12.09 &10.77 &2.19 &2.49 &1.91 &1.30 &0.84 &0.41 &0.22 &0.07 &0.01 &0.07 \\\hline
 \rowcolor{purple}\multirow{6}{*}{\rotatebox{90}{\textcolor{purple}{$\blacksquare$} DREAM}} & \textbf{Clean} & \textbf{44.66}&\textbf{57.07}&\textbf{63.25}&\textbf{67.75}&\textbf{68.96}&\textbf{24.95} &\textbf{36.17} &\textbf{42.36} &\textbf{44.46} &\textbf{46.84} &\textbf{7.59} &\textbf{17.58} &\textbf{18.88} &\textbf{19.91} &\textbf{17.46} \\
 & FGSM & 10.35 &7.86 &7.54 &5.48 &5.91 &4.36 &2.51 &2.70 &1.60 &1.61 &0.18 &0.20 &0.14 &0.03 &0.14 \\
 & PGD & 8.00 &4.40 &4.01 &2.37 &2.73 &3.10 &1.10 &1.05 &0.54 &0.44 &0.08 &0.01 &0.01 &0.00 &0.02 \\
 & CW & 7.79 &4.51 &4.23 &2.42 &2.79 &2.63 &1.14 &1.13 &0.60 &0.44 &0.07 &0.03 &0.00 &0.00 &0.01 \\
 & VMI & 8.38 &4.75 &4.23 &2.52 &2.95 &3.26 &1.17 &1.15 &0.55 &0.46 &0.09 &0.01 &0.03 &0.00 &0.02 \\
 & Jitter & 11.29 &9.66 &9.93 &9.22 &9.22 &2.99 &1.44 &1.57 &1.09 &0.95 &0.07 &0.03 &0.00 &0.00 &0.07 \\\hline

 \rowcolor{cyan}\multirow{6}{*}{\rotatebox{90}{\textcolor{cyan}{$\blacksquare$} D4M}} & \textbf{Clean} & \textbf{23.39}&\textbf{42.34}&\textbf{48.16}&\textbf{63.53}&\textbf{69.81}&\textbf{11.27} &\textbf{31.90} &\textbf{40.12} &\textbf{48.15} &\textbf{51.10} &\textbf{2.16} &\textbf{6.94} &\textbf{14.25} &\textbf{35.80} &\textbf{42.89} \\
 & FGSM & 3.62 &6.19 &9.05 &8.26 &8.50 &1.25 &1.15 &1.07 &1.12 &0.87 &0.29 &0.50 &0.53 &1.18 &1.83 \\
 & PGD & 2.44 &3.62 &5.86 &4.02 &3.84 &0.78 &0.35 &0.36 &0.29 &0.13 &0.23 &0.22 &0.16 &0.13 &0.30 \\
 & CW & 2.44 &3.94 &6.48 &4.46 &4.26 &0.85 &0.43 &0.41 &0.35 &0.17 &0.20 &0.28 &0.26 &0.24 &0.44 \\
 & VMI & 2.60 &3.93 &6.27 &4.42 &4.25 &0.82 &0.41 &0.36 &0.30 &0.14 &0.24 &0.25 &0.18 &0.16 &0.34\\
 & Jitter & 3.08 &5.43 &8.32 &8.78 &9.60 &0.99 &0.88 &0.66 &0.77 &0.52 &0.20 &0.26 &0.22 &0.35 &0.50 \\\hline
 

\multicolumn{2}{c|}{\multirow{4}{*}{\textbf{Whole}}} 
& &\textbf{Clean}&\textbf{FGSM}&\textbf{PGD}& & &\textbf{Clean}&\textbf{FGSM}&\textbf{PGD}& & &\textbf{Clean}&\textbf{FGSM}&\textbf{PGD}& \\& 
& &84.45&3.18&1.10& & &55.74&1.31&3.40& & &38.77&0.69&0.21& \\&

& &\textbf{CW}&\textbf{VMI}&\textbf{Jitter}& & &\textbf{CW}&\textbf{VMI}&\textbf{Jitter}& & &\textbf{CW}&\textbf{VMI}&\textbf{Jitter}& \\& 
& & 0.78 & 0.77 &9.04& & & 0.05 & 0.04 &0.38& & & 0.12 & 0.22 &0.17& \\
 \bottomrule
\end{tabular}
}
\end{table*}

\begin{table*}[p]
  \caption{Clean and robust accuracies of models trained on distilled CIFAR-10, CIFAR-100 and TinyImageNet. The perturbation budget is set to $|\varepsilon|=8/255$.}
\label{table:benchmark8/255}
\setlength{\tabcolsep}{2.5pt} 
\tiny
\renewcommand{\arraystretch}{1} 
\centering
\resizebox{\textwidth}{!}{
\begin{tabular}{cc|ccccc|ccccc|ccccc}
\toprule
\multicolumn{2}{c|}{Dataset} &\multicolumn{5}{c|}{\textbf{CIFAR-10}} & \multicolumn{5}{c|}{\textbf{CIFAR-100}} & \multicolumn{5}{c}{\textbf{TinyImageNet}} \\
\hline
\multicolumn{2}{c|}{IPC}   & 1 & 5 & 10 & 30 & 50 & 1 & 5 & 10 & 30 & 50 & 1 & 5 & 10 & 30 & 50 \\ \specialrule{0.8pt}{0pt}{0pt}
\rowcolor{nbarrier}\multirow{6}{*}{\rotatebox{90}{\textcolor{nbarrier}{$\blacksquare$} DC}} & \textbf{Clean} & \textbf{29.73} & \textbf{41.23} & \textbf{46.07} & \textbf{53.65} & \textbf{55.08} & \textbf{12.82} & \textbf{21.97} & \textbf{25.66} & \textbf{30.82} & \textbf{30.19} & \textbf{5.49} & \textbf{9.70} & \textbf{11.74} & \textbf{13.51} & \textbf{11.97} \\
 & FGSM & 5.76 &2.12 &0.93 &0.56 &0.36 &1.19 &2.07 &1.86 &0.48 &0.34 &0.03 &0.07 &0.03 &0.00 &0.01 \\
 & PGD & 4.21 &0.63 &0.19 &0.81 &0.01 &0.61 &0.92 &0.70 &0.11 &0.05 &0.01 &0.00 &0.00 &0.00 &0.00 \\
 & CW & 3.32 &0.54 &0.14 &0.08 &0.00 &0.15 &0.75 &0.66 &0.05 &0.01 &0.00 &0.00 &0.00 &0.00 &0.00 \\
 & VMI &4.38 &0.75 &0.25 &0.10 &0.03 &0.63 &0.99 &0.77 &0.13 &0.05 &0.01 &0.00 &0.00 &0.00 &0.00 \\
 & Jitter & 6.98 &6.34 &5.56 &6.15 &5.24 &0.44 &1.17 &1.07 &0.32 &0.40 &0.00 &0.00 &0.02 &0.01 &0.01 \\\hline
\rowcolor{nbicycle}\multirow{6}{*}{\rotatebox{90}{\textcolor{nbicycle}{$\blacksquare$} DSA}} & \textbf{Clean} & \textbf{29.27} & \textbf{48.20} & \textbf{52.93} & \textbf{56.41} & \textbf{61.14} & \textbf{14.38} & \textbf{27.13} & \textbf{32.94} & \textbf{37.29} & \textbf{42.87} & \textbf{5.47} & \textbf{13.99} & \textbf{17.47} & \textbf{17.12} & \textbf{21.89} \\
 & FGSM & 3.31 &1.24 &0.86 &0.23 &0.22 &1.18 &1.82 &1.33 &0.46 &0.93 &0.08 &0.09 &0.07 &0.00 &0.00 \\
 & PGD & 1.95 &0.26 &0.07 &0.02 &0.04 &0.57 &0.55 &0.27 &0.10 &0.37 &0.01 &0.00 &0.00 &0.00 &0.01 \\
 & CW & 2.28 &0.28 &0.08 &0.02 &0.00 &0.34 &0.57 &0.29 &0.02 &0.13 &0.01 &0.00 &0.00 &0.00 &0.00 \\
 & VMI & 2.26 &0.38 &0.11 &0.02 &0.01 &0.73 &0.72 &0.30 &0.04 &0.16 &0.01 &0.00 &0.00 &0.00 &0.00 \\
 & Jitter & 6.05 &6.83 &6.45 &4.57 &5.67 &0.45 &0.99 &0.92 &0.72 &0.79 &0.04 &0.02 &0.09 &0.02 &0.01 \\\hline
\rowcolor{nbus}\multirow{6}{*}{\rotatebox{90}{\textcolor{nbus}{$\blacksquare$} DM}} & \textbf{Clean} & \textbf{26.75} & \textbf{42.78} & \textbf{49.81} & \textbf{59.97} & \textbf{63.12} & \textbf{11.71} & \textbf{23.90} & \textbf{29.92} & \textbf{38.48} & \textbf{43.56} & \textbf{4.06} & \textbf{9.76} & \textbf{14.16} & \textbf{21.03} & \textbf{21.29} \\
 & FGSM & 5.83 &0.74 &0.50 &1.30 &0.71 &1.32 &0.62 &0.74 &0.83 &0.46 &0.11 &0.02 &0.03 &0.01 &0.01 \\
 & PGD & 4.44 &0.22 &0.10 &0.30 &0.35 &0.67 &0.08 &0.10 &0.10 &0.18 &0.05 &0.00 &0.00 &0.00 &0.00 \\
 & CW & 4.66 &0.27 &0.10 &0.19 &0.07 &0.53 &0.05 &0.14 &0.03 &0.01 &0.02 &0.00 &0.00 &0.00 &0.00 \\
 & VMI & 4.73 &0.28 &0.11 &0.17 &0.08 &0.75 &0.12 &0.13 &0.04 &0.04 &0.05 &0.00 &0.00 &0.00 &0.00 \\
 & Jitter & 6.52 &5.06 &4.66 &6.73 &7.64 &0.64 &0.18 &0.32 &0.24 &0.34 &0.03 &0.00 &0.00 &0.02 &0.02 \\\hline
\rowcolor{ncar}\multirow{6}{*}{\rotatebox{90}{\textcolor{ncar}{$\blacksquare$} MTT}} & \textbf{Clean} & \textbf{45.74} & \textbf{57.19} & \textbf{60.98} & \textbf{66.23} & \textbf{70.45} & \textbf{20.30} & \textbf{34.81} & \textbf{37.84} & \textbf{42.65} & \textbf{44.34} & \textbf{8.91} & \textbf{14.86} & \textbf{19.93} & \textbf{23.05} & \textbf{26.16} \\
 & FGSM & 1.38 &0.56 &0.60 &0.32 &0.15 &1.10 &0.91 &0.45 &0.43 &0.52 &0.03 &0.00 &0.01 &0.00 &0.00 \\
 & PGD & 0.38 &0.09 &0.08 &0.02 &0.00 &0.57 &0.16 &0.03 &0.02 &0.02 &0.01 &0.00 &0.00 &0.00 &0.00 \\
 & CW & 0.21 &0.05 &0.02 &0.01 &0.00 &0.17 &0.11 &0.03 &0.01 &0.00 &0.00 &0.00 &0.00 &0.00 &0.00 \\
 & VMI & 0.49 &0.13 &0.11 &0.03 &0.00 &0.62 &0.18 &0.04 &0.03 &0.04 &0.01 &0.00 &0.00 &0.00 &0.00 \\
 & Jitter & 5.92 &6.39 &6.53 &6.89 &6.84 &0.40 &0.58 &0.48 &0.44 &0.56 &0.02 &0.00 &0.02 &0.03 &0.02 \\\hline
\rowcolor{npedestrian}\multirow{6}{*}{\rotatebox{90}{\textcolor{npedestrian}{$\blacksquare$} TESLA}} & \textbf{Clean} & \textbf{47.07} & \textbf{57.23} & \textbf{61.78} & \textbf{67.08} & \textbf{68.83} & \textbf{20.00} & \textbf{31.65} & \textbf{34.99} & \textbf{39.16} & \textbf{45.92} & \textbf{7.40} & \textbf{16.01} & \textbf{19.45} & \textbf{23.71} & \textbf{26.63} \\
 & FGSM & 1.30 &0.27 &0.12 &0.26 &0.18 &0.96 &0.93 &0.23 &0.41 &0.49 &0.00 &0.00 &0.00 &0.00 &0.00 \\
 & PGD & 0.42 &0.02 &0.01 &0.02 &0.03 &0.52 &0.17 &0.00 &0.01 &0.00 &0.00 &0.00 &0.00 &0.00 &0.00 \\
 & CW & 0.37 &0.01 &0.01 &0.02 &0.00 &0.32 &0.11 &0.00 &0.00 &0.02 &0.00 &0.00 &0.00 &0.00 &0.00 \\
 & VMI & 0.56 &0.02 &0.01 &0.05 &0.01 &0.58 &0.23 &0.01 &0.01 &0.02 &0.00 &0.00 &0.00 &0.00 &0.00 \\
 & Jitter & 6.21 &6.49 &5.80 &6.82 &6.40 &0.59 &0.40 &0.24 &0.30 &0.49 &0.01 &0.02 &0.02 &0.04 &0.02 \\\hline
\rowcolor{nterrain}\multirow{6}{*}{\rotatebox{90}{\textcolor{nterrain}{$\blacksquare$} SRe2L}} & \textbf{Clean} & \textbf{13.49} & \textbf{31.06} & \textbf{37.53} & \textbf{54.88} & \textbf{63.28} & \textbf{4.68} & \textbf{28.92} & \textbf{39.64} & \textbf{51.33} & \textbf{53.95} & \textbf{6.28} & \textbf{18.38} & \textbf{26.92} & \textbf{39.49} & \textbf{43.24} \\
 & FGSM & 2.94 &1.35 &0.95 &1.23 &1.61 &0.93 &1.86 &2.92 &3.42 &3.88 &0.22 &0.12 &0.18 &0.29 &0.37 \\
 & PGD & 2.30 &0.32 &0.11 &0.01 &0.01 &0.61 &0.10 &0.02 &0.02 &0.00 &0.09 &0.00 &0.00 &0.02 &0.00 \\
 & CW & 2.19 &0.23 &0.09 &0.03 &0.05 &0.53 &0.01 &0.00 &0.00 &0.01 &0.10 &0.00 &0.00 &0.00 &0.00 \\
 & VMI & 2.31 &0.46 &0.19 &0.03 &0.03 &0.62 &0.10 &0.03 &0.05 &0.00 &0.11 &0.00 &0.00 &0.00 &0.00 \\
 & Jitter & 2.70 &7.24 &7.38 &8.42 &9.89 &0.60 &0.37 &1.29 &1.87 &2.00 &0.09 &0.01 &0.02 &0.07 &0.12 \\\hline
 \rowcolor{grey}\multirow{6}{*}{\rotatebox{90}{\textcolor{grey}{$\blacksquare$} IDM}} & \textbf{Clean} & \textbf{46.14} & \textbf{54.45} & \textbf{58.84} & \textbf{65.83} & \textbf{67.82} & \textbf{23.90} & \textbf{40.02} & \textbf{45.33} & \textbf{45.83} & \textbf{46.13} & \textbf{10.42} & \textbf{22.39} & \textbf{22.4} & \textbf{24.96} & \textbf{26.89} \\
 & FGSM & 1.67 &2.32 &1.10 &0.73 &0.66 &0.88 &0.93 &0.82 &0.62 &0.39 &0.11 &0.02 &0.03 &0.02 &0.01 \\
 & PGD & 0.60 &0.76 &0.38 &2.15 &2.53 &0.27 &0.39 &0.44 &0.58 &0.41 &0.03 &0.10 &0.07 &0.03 &0.02 \\
 & CW & 0.60 &0.73 &0.22 &0.05 &0.04 &0.22 &0.13 &0.09 &0.01 &0.00 &0.02 &0.00 &0.00 &0.00 &0.00 \\
 & VMI & 0.76 &0.92 &0.27 &0.08 &0.04 &0.34 &0.17 &0.07 &0.02 &0.01 &0.04 &0.00 &0.00 &0.00 &0.00 \\
 & Jitter & 7.92 &9.50 &8.30 &8.61 &8.10 &0.41 &0.52 &0.37 &0.31 &0.20 &0.04 &0.00 &0.00 &0.02 &0.02 \\\hline
 \rowcolor{purple}\multirow{6}{*}{\rotatebox{90}{\textcolor{purple}{$\blacksquare$} DREAM}}& \textbf{Clean} & \textbf{44.66}&\textbf{57.07}&\textbf{63.25}&\textbf{67.75}&\textbf{68.96}&\textbf{24.95} &\textbf{36.17} &\textbf{42.36} &\textbf{44.46} &\textbf{46.84} &\textbf{7.59} &\textbf{17.58} &\textbf{18.88} &\textbf{19.91} &\textbf{17.46} \\
 & FGSM & 2.12 &0.89 &0.64 &0.34 &0.38 &1.12 &0.49 &0.53 &0.30 &24.00 &0.03 &0.00 &0.00 &0.00 &0.00 \\
 & PGD & 0.81 &0.09 &0.05 &0.03 &0.01 &0.35 &0.05 &0.02 &0.00 &0.00 &0.00 &0.00 &0.00 &0.00 &0.00 \\
 & CW & 0.83 &0.10 &0.06 &0.02 &0.01 &0.32 &0.05 &0.02 &0.00 &0.00 &0.00 &0.00 &0.00 &0.00 &0.00 \\
 & VMI & 1.01 &0.13 &0.09 &0.04 &0.04 &0.42 &0.05 &0.04 &0.00 &0.00 &0.00 &0.00 &0.00 &0.00 &0.00 \\
 & Jitter & 7.40 &7.08 &7.80 &6.54 &7.07 &0.63 &0.45 &0.52 &0.49 &0.39 &0.01 &0.02 &0.02 &0.00 &0.02 \\\hline

 \rowcolor{cyan}\multirow{6}{*}{\rotatebox{90}{\textcolor{cyan}{$\blacksquare$} D4M}} & \textbf{Clean} & \textbf{23.39}&\textbf{42.34}&\textbf{48.16}&\textbf{63.53}&\textbf{69.81}&\textbf{11.27} &\textbf{31.90} &\textbf{40.12} &\textbf{48.15} &\textbf{51.10} &\textbf{2.16} &\textbf{6.94} &\textbf{14.25} &\textbf{35.80} &\textbf{42.89} \\
 & FGSM & 0.63 &1.05 &1.75 &0.82 &0.86 &0.18 &0.08 &0.16 &0.22 &0.23 &0.16 &0.18 &0.19 &0.38 &0.59 \\
 & PGD & 0.07 &0.18 &0.33 &0.10 &0.06 &0.05 &0.01 &0.00 &0.00 &0.00 &0.07 &0.05 &0.01 &0.00 &0.02 \\
 & CW & 0.15 &0.18 &0.34 &0.11 &0.08 &0.04 &0.01 &0.00 &0.00 &0.00 &0.09 &0.06 &0.00 &0.00 &0.02 \\
 & VMI & 0.14 &0.20 &0.42 &0.13 &0.08 &0.05 &0.01 &0.00 &0.00 &0.00 &0.09 &0.05 &0.01 &0.00 &0.02 \\
 & Jitter & 1.35 &2.88 &3.86 &5.23 &5.96 &0.07 &0.06 &0.08 &0.21 &0.21 &0.07 &0.05 &0.04 &0.11 &0.15 \\\hline
 

\multicolumn{2}{c|}{\multirow{4}{*}{\textbf{Whole}}} 
& &\textbf{Clean}&\textbf{FGSM}&\textbf{PGD}& & &\textbf{Clean}&\textbf{FGSM}&\textbf{PGD}& & &\textbf{Clean}&\textbf{FGSM}&\textbf{PGD}& \\& 
& &84.45&0.12&0.57& & &55.74&0.76&2.83& & &38.77&0.04&0.00& \\&

& &\textbf{CW}&\textbf{VMI}&\textbf{Jitter}& & &\textbf{CW}&\textbf{VMI}&\textbf{Jitter}& & &\textbf{CW}&\textbf{VMI}&\textbf{Jitter}& \\& 
& & 0.00 & 0.00 &8.06& & & 0.00 & 0.00 &0.32& & & 0.00 & 0.00 &0.07& \\
 \bottomrule
\end{tabular}
}
\end{table*}

\begin{table*}[t]
  \caption{Original accuracies and robust accuracies of models trained on ImageNet-subsets. MTT is applied in the distillation stage. The perturbation budget is set to $\left| \varepsilon \right|=4/255$.}\label{table:imagenetsubset4/255}
\centering
\setlength{\tabcolsep}{2.5pt} 
\renewcommand{\arraystretch}{1.1} 
\resizebox{1\textwidth}{!}{
\begin{tabular}{c|cccccc|cccccc|cccccc}
\toprule
      \multirow{2}{*}{IPC}               & \multicolumn{6}{c|}{\cellcolor{nbarrier}\textbf{ImageNette}}        & \multicolumn{6}{c|}{\cellcolor{nbicycle}\textbf{ImageWoof}}         & \multicolumn{6}{c}{\cellcolor{nbus}\textbf{ImageFruit}}        \\
 & \textbf{Clean} & FGSM  & PGD   & CW& VMI&Jitter    & \textbf{Clean} & FGSM  & PGD   & CW  & VMI&Jitter  & \textbf{Clean} & FGSM  & PGD   & CW  & VMI&Jitter   \\
 \hline
1 & \textbf{48.20} & 7.40 & 5.40 & 4.80 & 5.60 & 8.00 & \textbf{30.40} & 2.60 & 2.00 & 1.40 & 2.00 & 2.20 & \textbf{25.00} & 2.80 & 1.80 & 1.60 & 1.80 & 2.60 \\
5 & \textbf{62.00} & 11.00 & 5.00 & 5.80 & 5.60 & 14.00 & \textbf{35.20} & 1.00 & 0.40 & 0.40 & 0.40 & 1.40 & \textbf{41.00} & 2.20 & 1.00 & 0.80 & 1.00 & 2.60 \\
10 & \textbf{66.40} & 10.80 & 4.60 & 4.60 & 5.40 & 12.20 & \textbf{38.00} & 0.80 & 0.20 & 0.40 & 0.40 & 2.40 & \textbf{42.20} & 2.40 & 0.40 & 0.60 & 0.60 & 3.20 \\
30 & \textbf{66.60} & 8.80 & 3.20 & 3.00 & 3.40 & 11.60 & \textbf{38.80} & 0.60 & 0.20 & 0.20 & 0.20 & 1.60 & \textbf{44.40} & 3.00 & 0.80 & 0.80 & 1.20 & 3.40 \\
50 & \textbf{67.60} & 8.40 & 2.60 & 1.40 & 2.00 & \textbf{13.00} & \textbf{39.40} & 0.00 & 0.00 & 0.00 & 0.00 & 0.40 & \textbf{44.60} & 2.20 & 1.00 & 1.00 & 1.00 & 3.60 \\
\rowcolor{grey!40}\textbf{Whole} & \textbf{86.40} & 6.60 & 4.40 & 1.00 & 1.20 & 11.20 & \textbf{67.20} & 0.00 & 0.00 & 0.00 & 0.00 & 1.20 & \textbf{67.40} & 2.60 & 0.60 & 0.40 & 0.40 & 4.00   \\ \toprule
      \multirow{2}{*}{IPC}                 & \multicolumn{6}{c|}{\cellcolor{ncar}\textbf{ImageMeow}}         & \multicolumn{6}{c|}{\cellcolor{npedestrian}\textbf{ImageSquawk}}       & \multicolumn{6}{c}{\cellcolor{nterrain}\textbf{ImageYellow}}       \\ 
 & \textbf{Clean} & FGSM  & PGD   & CW & VMI&Jitter  & \textbf{Clean} & FGSM  & PGD   & CW & VMI&Jitter     & \textbf{Clean} & FGSM  & PGD   & CW  & VMI&Jitter    \\

\hline
1 & \textbf{31.00} & 2.00 & 0.60 & 0.20 & 0.80 & 1.60 & \textbf{39.00} & 4.20 & 1.80 & 2.20 & 2.40 & 5.60 & \textbf{44.60} & 6.40 & 3.00 & 2.60 & 4.00 & 7.40 \\
5 & \textbf{41.40} & 0.80 & 0.20 & 0.40 & 0.40 & 3.00 & \textbf{52.40} & 6.60 & 3.00 & 2.40 & 3.40 & 6.60 & \textbf{59.20} & 8.40 & 4.60 & 4.80 & 4.60 & 13.60 \\
10 & \textbf{44.40} & 1.40 & 0.80 & 0.60 & 0.80 & 2.20 & \textbf{55.60} & 4.40 & 2.40 & 2.60 & 2.80 & 6.40 & \textbf{63.40} & 9.40 & 5.00 & 5.60 & 5.80 & 11.40 \\
30 & \textbf{44.20} & 0.80 & 0.60 & 0.60 & 0.60 & 3.40 & \textbf{56.20} & 4.00 & 1.60 & 2.00 & 2.00 & \textbf{6.00} & 65.40 & 7.80 & 4.20 & 4.40 & 3.80 & 10.00 \\
50 & \textbf{44.20} & 0.80 & 0.60 & 0.60 & 0.60 & 2.60 & \textbf{59.20} & 3.80 & 2.00 & 1.60 & 1.60 & 7.00 & \textbf{66.20} & 9.00 & 4.60 & 3.80 & 3.80 & 10.40 \\
\rowcolor{grey!40}\textbf{Whole} & \textbf{69.00} & 0.40 & 0.40 & 0.40 & 0.40 & 1.60 & \textbf{86.40} & 7.20 & 4.40 & 3.00 & 2.80 & 13.60 & \textbf{84.40} & 8.00 & 9.20 & 2.20 & 2.00 & 13.00 \\ \bottomrule
\end{tabular}}
\end{table*}

\begin{table*}[t]
  \caption{Original accuracies and robust accuracies of models trained on ImageNet-subsets. MTT is applied in the distillation stage. The perturbation budget is set to $\left| \varepsilon \right|=8/255$.}\label{table:imagenetsubset8/255}
\centering
\setlength{\tabcolsep}{2.5pt} 
\renewcommand{\arraystretch}{1.1} 
\resizebox{1\textwidth}{!}{
\begin{tabular}{c|cccccc|cccccc|cccccc}
\toprule
      \multirow{2}{*}{IPC}               & \multicolumn{6}{c|}{\cellcolor{nbarrier}\textbf{ImageNette}}        & \multicolumn{6}{c|}{\cellcolor{nbicycle}\textbf{ImageWoof}}         & \multicolumn{6}{c}{\cellcolor{nbus}\textbf{ImageFruit}}        \\
 & \textbf{Clean} & FGSM  & PGD   & CW& VMI&Jitter    & \textbf{Clean} & FGSM  & PGD   & CW  & VMI&Jitter  & \textbf{Clean} & FGSM  & PGD   & CW  & VMI&Jitter   \\
 \hline
1 & \textbf{48.20} & 1.20 & 0.20 & 0.20 & 0.40 & 6.60 & \textbf{30.40} & 0.60 & 0.40 & 0.20 & 0.40 & 1.80 & \textbf{25.00} & 0.80 & 0.00 & 0.00 & 0.00 & 2.20 \\
5 & \textbf{62.00} & 1.40 & 0.20 & 0.20 & 0.20 & 9.40 & \textbf{35.20} & 0.00 & 0.00 & 0.00 & 0.00 & 1.60 & \textbf{41.00} & 0.20 & 0.00 & 0.00 & 0.00 & 1.20 \\
10 & \textbf{66.40} & 0.80 & 0.20 & 0.20 & 0.20 & 11.40 & \textbf{38.00} & 0.00 & 0.00 & 0.00 & 0.00 & 2.60 & \textbf{42.20} & 0.20 & 0.00 & 0.00 & 0.00 & 2.40 \\
30 & \textbf{66.60} & 1.80 & 1.20 & 0.00 & 0.00 & 7.80 & \textbf{38.80} & 0.00 & 0.00 & 0.00 & 0.00 & 2.60 & \textbf{44.40} & 0.20 & 0.00 & 0.00 & 0.00 & 2.40 \\
50 & \textbf{67.60} & 1.80 & 1.20 & 0.20 & 0.20 & 9.80 & \textbf{39.40} & 0.00 & 0.00 & 0.00 & 0.00 & 2.00 & \textbf{44.60} & 0.40 & 0.20 & 0.00 & 0.00 & 3.20 \\
\rowcolor{grey!40}\textbf{Whole} & \textbf{86.40} & 0.40 & 3.60 & 0.00 & 0.00 & \textbf{7.20} & \textbf{67.20} & 0.00 & 0.00 & 0.00 & 0.00 & \textbf{0.00} & \textbf{67.40} & 0.00 & 0.60 & 0.00 & 0.00 & 2.60   \\ 
\toprule
      \multirow{2}{*}{IPC}                 & \multicolumn{6}{c|}{\cellcolor{ncar}\textbf{ImageMeow}}         & \multicolumn{6}{c|}{\cellcolor{npedestrian}\textbf{ImageSquawk}}       & \multicolumn{6}{c}{\cellcolor{nterrain}\textbf{ImageYellow}}       \\ 
 & \textbf{Clean} & FGSM  & PGD   & CW & VMI&Jitter  & \textbf{Clean} & FGSM  & PGD   & CW & VMI&Jitter     & \textbf{Clean} & FGSM  & PGD   & CW  & VMI&Jitter    \\

\hline
1 & \textbf{31.00} & 0.00 & 0.00 & 0.00 & 0.00 & 2.60 & \textbf{39.00} & 0.40 & 0.00 & 0.20 & 0.00 & 3.80 & \textbf{44.60} & 0.60 & 0.00 & 0.00 & 0.00 & 6.00 \\
5 & \textbf{41.40} & 0.00 & 0.00 & 0.00 & 0.00 & 1.80 & \textbf{52.40} & 0.20 & 0.00 & 0.00 & 0.00 & 6.60 & \textbf{59.20} & 2.20 & 0.60 & 0.20 & 0.60 & 7.20 \\
10 & \textbf{44.40} & 0.40 & 0.20 & 0.00 & 0.00 & 1.60 & \textbf{55.60} & 0.60 & 0.00 & 0.20 & 0.20 & 5.40 & \textbf{63.40} & 2.00 & 0.80 & 0.40 & 1.00 & 9.80 \\
30 & \textbf{44.20} & 0.00 & 0.00 & 0.00 & 0.00 & 1.00 & \textbf{56.20} & 0.40 & 0.40 & 0.20 & 0.20 & 5.60 & \textbf{65.40} & 2.00 & 2.00 & 0.20 & 0.40 & 10.80 \\
50 & \textbf{44.20} & 0.20 & 0.20 & 0.00 & 0.00 & 2.40 & \textbf{59.20} & 1.00 & 1.40 & 0.20 & 0.20 & 4.60 & \textbf{66.20} & 1.80 & 2.20 & 0.00 & 0.00 & 8.80 \\
\rowcolor{grey!40}\textbf{Whole} & \textbf{69.00} & 0.00 & 0.20 & 0.00 & 0.00 & 2.40 & \textbf{86.40} & 1.40 & 3.60 & 0.00 & 0.00 & 7.60 & \textbf{84.40} & 0.40 & 8.20 & 0.00 & 0.00 & 7.20 \\ \bottomrule
\end{tabular}}
\end{table*}

\vspace{-0.75em}
\subsection{Component Analysis}
\label{sec:component}

In the previous section, we evaluated the robustness of a wide range of distillation methods. These methods encompass various optimization objectives, including gradient matching, distribution matching, trajectory matching, optimization-based approaches and generation-based approaches.
Additionally, we observed that many advanced methods enhance clean accuracy by incorporating certain heuristic components into the distillation process.
For example, DSA \cite{Zhao:DSA} incorporates various data augmentations in both distillation and training process based on DC \cite{Zhao:DC}. IDC~\cite{kim2022idc} has the same optimization objective with DC while the main difference between them lies in the inclusion of multi-formation downsampling operation.
DREAM \cite{Liu:DREAM}, while based on IDC, distinguishes itself by integrating clustering operations, such as K-means.

Since such components are crucial for dataset distillation, analyzing their impact on robustness would offer an opportunity to have further insights into distillation process. To this end, we designed three sets of experiments to investigate the impact of components involved in distillation process on robustness, including data augmentation operations, downsampling factor and clustering number. 
These experiments were conducted on CIFAR-10 with an IPC of $10$, keeping all other parameters consistent.

We first investigate the impact of data augmentation on DSA. In both distillation and training stage, the DSA strategy \cite{Zhao:DSA} randomly selects one operation each time from a series of data augmentations including color jitter, crop, cutout, flip, rotate and scale. We individually selected each of these augmentations and applied them to both the distillation and training stages. Results are shown in Table \ref{tab:aug}. We found that only using the flip operation achieves the best robustness, even surpassing the robust accuracy obtained with the full DSA strategies, while crop and cutout shows relatively lower robustness. This indicates that although rich data augmentation can enhance the model's clean accuracy, certain augmentations may undermine model robustness.

In the case of IDC \cite{kim2022idc}, the approach employs multi-formation functions to generate synthetic data with lower resolution.
Specifically, larger downsampling factor $f$ results in more synthetic data while causing lower resolution of each one accordingly. We evaluate different fraction factors and show them in Table \ref{tab:downsampling}. We found that a higher fraction of distilled images leads to worse robustness.
We hypothesized that an excessive downsampling ratio may lead to the loss of critical image information, adversely affecting the model’s robustness.

Lastly, DREAM \cite{Liu:DREAM} improves the sample selection in distillation by involving clustering operation, which leads to higher evaluation accuracy with certain clustering number. We select different clustering number $N$ for exploring the relation between selected samples and robustness. As shown in Table \ref{tab:cluster}, while the robust accuracy is slightly better when $N=64$ compared to other settings, the overall robustness is not significantly affected by the clustering operation. This may indicates that, although clustering operation selects samples that are more effective and efficient to improving accuracy as the matching targets, it has weak influence on the extraction of robust and non-robust feature.

Our experimental results indicate that while many advanced methods improve clean accuracy by adding various components to the distillation process, some of them may negatively affect model robustness. 
Therefore, in addition to achieving the highest possible clean accuracy, it is also crucial to consider how to adjust the corresponding hyperparameters to balance the models' adversarial robustness.

\vspace{-0.75em}
\section{Conclusion}
\label{sec:conclusion}
In this paper, we present a comprehensive benchmark for evaluating the adversarial robustness of distilled datasets, overcoming prior limitations. 
Our study integrates the latest distillation methods, diverse adversarial attacks, and large-scale datasets. 
We find that distilled datasets generally exhibit better robustness than original datasets, with robustness declining as IPC increases. 
Additionally, we delve deeper into the impact of different components used during distillation process on robustness, including data augmentation, downsampling factor and clustering number. Our quantitative analysis indicates that clustering operation has weak infulence on robustness while some data augmentations like crop and cutout would cause worse robustness. We also find that robust accuracy drops with higher fraction factor of downsampling operation.
Our work offers a new evaluation perspective for dataset distillation and suggests potential research directions.

\vspace{-1em}
\bibliographystyle{IEEEtran}
\bibliography{IEEEabrv,main.bib}

\newpage
\clearpage
\section*{Supplemental Materials}
\subsection{Outlook}
In the \textit{Experimental Results} and \textit{Component Analysis} sections, we present the robustness of the distilled dataset under various configurations, along with analysis on components involved in distillation process.  We believe our exploratory research and publicly available benchmarks would provide directions for enhancing dataset distillation in future works and offer new perspectives for evaluating their efficacy.

One potential aspect to consider is how to strike a balance between robustness and accuracy. While previous studies \cite{tsipras2019robustness} have demonstrated a trade-off between robustness and accuracy, these two factors are further coupled with IPC in dataset distillation. When the IPC is too restrictive, overfitting would lead to poor performances. On the other hand, simply increasing the training data can cause a decrease in the model's ability to withstand adversarial attacks. We emphasize the need for future research to fully acknowledge and address this challenge in dataset distillation.

\subsection{Implement Details for MTT and TESLA}
\label{sec:a-details}
The authors of MTT \cite{Cazenavette:MTT} and TESLA \cite{cui2023scaling} applied the ZCA whitening \cite{kessy2018optimal} for accelerating convergence under several IPC settings. In our experiments, we discarded it and did some modification to distillation parameters including learning rate, synthetic steps, etc. We show the parameters of our distillation in Table \ref{tab:a-mtt} and Table \ref{tab:a-tesla}.

\subsection{Cross-architecture experiments}
\label{sec:cross}
\cite{Guo:Lossless} and \cite{Cui:dcbench} both suggest that deviating from ConvNet to other network architectures results in notable performance degradation. 
\cite{Cui:dcbench} highlights that almost all representative distillation methods opt for ConvNet for the distillation process.
Based on these observations, we selected the network architecture proposed by the original authors in our work.
To offer additional insights, we trained three more different network architectures on the distilled datasets and evaluated their robustness. 
The datasets are distilled with ConvNet. 
Their clean and robust accuracy results are shown in Table \ref{tab:a-cross}.
It is worth noting that small distilled datasets can be overly simplistic for networks like ResNet18, particularly in the case of DC, where no data augmentation is applied. 
This simplicity can lead to overfitting, significantly compromising adversarial robustness. 
Similar to the evaluation results of ConvNet, these three networks also show a trend where robustness decreases as IPC increases. 
We also observe that ResNet18 and VGG11 shows relatively lower robust on MTT, while MobileNet behaves lower robust on DSA.

\subsection{Benchmark in Practice}
\label{sec:a-practice}
In the main text we illustrate our evaluation pipeline in Figure 1. The implementation details of our pipeline can be found at \href{https://github.com/FredWU-HUST/DD-RobustBench}{DD-RobustBench}. 
Below, we provide essential instructions for deploying and utilizing our benchmark.
\begin{itemize}
\item \textbf{Installation and Environment Setup} To ensure the benchmark operates correctly, please configure your Python environment according to the following specifications:
\begin{verbatim}
Python >= 3.11  Numpy >= 1.25.2
Torch >= 2.0.1  Torchattacks >= 3.4.0
Torchvision >= 0.15.2  Scipy >= 1.11.1
\end{verbatim}

\item \textbf{Model Training} To train a model on a distilled dataset, execute the script below. This will train a randomly initialized network and save the results accordingly:
\begin{verbatim}
python train.py --dataset CIFAR10
--model ConvNet --model_num 5
--train_batch 256  --train_epoch 1000
--save_path ./result/convnet_cifar10_dc
--optimizer sgd --distill_method DC 
--pt_path <path_to_distilled_dataset>
 --distilled [--dsa]
\end{verbatim}

\item \textbf{Evaluation} To evaluate the trained models on clean original test sets:
\begin{verbatim}
python eval.py --dataset CIFAR10
--model ConvNet --data_path ./data/
--pt_path <path_to_weight>
\end{verbatim}

\item \textbf{Attack Testing} To assess the robustness of the trained models under adversarial conditions, utilize the following script. 
The {\tt{weights.yaml}} contains weight files of the trained models. The {\tt{params.yaml}} contains parameters for different attacking algorithms. You can refer to the example files for more information. By running this, the users would acquire the adversarial robust accuracy of trained models and attacks specified.
\begin{verbatim}
python robust.py --dataset CIFAR10
--model ConvNet --attacker FGSM
--log_path <path_to_save_output>
--weights <path_to_weights.yaml>
--params <path_to_params.yaml>
--repeat 5  
\end{verbatim}
By running this, users would acquire the adversarial robust accuracy of trained models and attacks specified:
\begin{verbatim}
Using FGSM for perturbation.
Repeat each experiment for 5 times.
Test parameters: [0.00784, 0.00784, 10]
Testing weights: cifar10_d4m_ipc1.pth
Average robust accuracy: 8.92%
\end{verbatim}

\item \textbf{Extending the Benchmark} 
To enhance the benchmark's capabilities, researchers are encouraged to integrate new distillation methods, attacking algorithms, and network architectures. 
New distillation methods can be incorporated by updating the \texttt{load\_distilled\_dataset} function within \texttt{datasets.py}. 
For those interested in adding new attacking algorithms, it is possible to create an \texttt{ATTACK} class in \texttt{attack\_utils.py} and modify the \texttt{perturb} method accordingly. 
Additionally, to introduce new network architectures, define the network structure within \texttt{models.py} and implement instantiation in the \texttt{get\_network} function. 
These extensions are designed to facilitate broader research and application, ensuring that the benchmark remains adaptable and comprehensive.
\end{itemize} 

We will continue to update our repository to satisfy more SOTA algorithms.

\begin{table*}[t]
  \caption{Parameters for MTT in our experiments. We discard the ZCA under all IPC settings.}
\label{tab:a-mtt}
\centering
\setlength{\tabcolsep}{3pt}
\resizebox{0.7\textwidth}{!}{
\begin{tabular}{ccccccccc}
\hline
\textbf{Dataset} & \textbf{IPC} & \begin{tabular}[c]{@{}c@{}}\textbf{Sythetic}\\ \textbf{Steps}\end{tabular} & \textbf{Expert Epochs} & \begin{tabular}[c]{@{}c@{}}\textbf{Max Start}\\ \textbf{Epoch}\end{tabular} & \textbf{Batch Size} & \textbf{Pixels LR} & \textbf{Step LR} & \begin{tabular}[c]{@{}c@{}}\textbf{Starting}\\ \textbf{Step}\end{tabular} \\ \hline
\multirow{5}{*}{CIFAR-10} & 1 & 50 & 2 & 5 & - & $10^{3}$ & $10^{-7}$ & $10^{-2}$ \\
 & 5 & 40 & 2 & 10 & - & $10^{3}$ & $10^{-6}$ & $10^{-2}$ \\
 & 10 & 30 & 2 & 15 & - & $10^{3}$ & $10^{-5}$ & $10^{-2}$ \\
 & 30 & 30 & 2 & 40 & - & $10^{3}$ & $10^{-5}$ & $10^{-3}$ \\
 & 50 & 30 & 2 & 40 & - & $10^{3}$ & $10^{-5}$ & $10^{-3}$ \\ \hline
\multirow{5}{*}{CIFAR-100} & 1 & 40 & 3 & 15 & - & $10^{3}$ & $10^{-5}$ & $10^{-2}$ \\
 & 5 & 30 & 2 & 30 & - & $10^{3}$ & $10^{-5}$ & $10^{-2}$ \\
 & 10 & 20 & 2 & 40 & - & $10^{3}$ & $10^{-5}$ & $10^{-2}$ \\
 & 30 & 40 & 2 & 40 & 125 & $10^{3}$ & $10^{-5}$ & $10^{-2}$ \\
 & 50 & 80 & 2 & 40 & 125 & $10^{3}$ & $10^{-5}$ & $10^{-2}$ \\ \hline
\multirow{5}{*}{TinyImageNet} & 1 & 10 & 2 & 10 & - & $10^{4}$ & $10^{-4}$ & $10^{-2}$ \\
 & 5 & 15 & 2 & 25 & 200 & $10^{4}$ & $10^{-4}$ & $10^{-2}$ \\
 & 10 & 20 & 2 & 40 & 200 & $10^{4}$ & $10^{-4}$ & $10^{-2}$ \\
 & 30 & 20 & 2 & 40 & 200 & $10^{4}$ & $10^{-4}$ & $10^{-2}$ \\
 & 50 & 20 & 2 & 40 & 300 & $10^{4}$ & $10^{-4}$ & $10^{-2}$ \\ \hline
\end{tabular}%
}
\end{table*}

\begin{table*}[t]
\caption{Parameters for TESLA in our experiments. We discard the ZCA under all IPC settings.}
\label{tab:a-tesla}
\centering
\setlength{\tabcolsep}{3pt}
\resizebox{0.7\textwidth}{!}{%
\begin{tabular}{ccccccccc}
\hline
\textbf{Dataset} & \textbf{IPC} & \begin{tabular}[c]{@{}c@{}}\textbf{Sythetic}\\ \textbf{Steps}\end{tabular} & \textbf{Expert} \textbf{Epochs} & \begin{tabular}[c]{@{}c@{}}\textbf{Max} \textbf{Start}\\ \textbf{Epoch}\end{tabular} & \textbf{Batch} \textbf{Size} & \textbf{Pixels} \textbf{LR} & \textbf{Step LR} & \begin{tabular}[c]{@{}c@{}}\textbf{Starting}\\ \textbf{Step}\end{tabular} \\ \hline
\multirow{5}{*}{CIFAR-10} & 1 & 50 & 2 & 5 & - & $10^{3}$ & $10^{-7}$ & $10^{-2}$ \\
 & 5 & 40 & 2 & 10 & - & $10^{3}$ & $10^{-6}$ & $10^{-2}$ \\
 & 10 & 30 & 2 & 20 & - & $10^{3}$ & $10^{-5}$ & $10^{-2}$ \\
 & 30 & 30 & 2 & 40 & - & $10^{3}$ & $10^{-5}$ & $10^{-3}$ \\
 & 50 & 30 & 3 & 40 & - & $10^{3}$ & $10^{-5}$ & $10^{-3}$ \\ \hline
\multirow{5}{*}{CIFAR-100} & 1 & 40 & 3 & 15 & - & $10^{3}$ & $10^{-5}$ & $10^{-2}$ \\
 & 5 & 40 & 2 & 10 & - & $10^{3}$ & $10^{-6}$ & $10^{-2}$ \\
 & 10 & 15 & 3 & 30 & - & $10^{3}$ & $10^{-5}$ & $10^{-2}$ \\
 & 30 & 40 & 2 & 20 & 125 & $10^{3}$ & $10^{-5}$ & $10^{-2}$ \\
 & 50 & 80 & 2 & 40 & 125 & $10^{3}$ & $10^{-5}$ & $10^{-2}$ \\ \hline
\multirow{5}{*}{TinyImageNet} & 1 & 10 & 2 & 10 & - & $10^{4}$ & $10^{-4}$ & $10^{-2}$ \\
 & 5 & 15 & 2 & 25 & 200 & $10^{4}$ & $10^{-4}$ & $10^{-2}$ \\
 & 10 & 20 & 2 & 40 & 200 & $10^{4}$ & $10^{-4}$ & $10^{-2}$ \\
 & 30 & 20 & 2 & 40 & 200 & $10^{4}$ & $10^{-4}$ & $10^{-2}$ \\
 & 50 & 20 & 2 & 40 & 300 & $10^{4}$ & $10^{-4}$ & $10^{-2}$ \\ \hline
\end{tabular}%
}
\end{table*}

\begin{table*}[]
\centering
\caption{Cross-architecture results on clean and perturbed test set. We trained $3$ different network architectures on distilled datasets. The datasets are distilled with ConvNet. Perturbation is fixed to $|\varepsilon|=2/255$.}
\label{tab:a-cross}
\tiny
\renewcommand{\arraystretch}{1.1}
\setlength{\tabcolsep}{3pt}
\resizebox{0.95\textwidth}{!}{%
\begin{tabular}{ccccccccccccccc}
\hline
\multirow{2}{*}{\textbf{Attack}} & \multirow{2}{*}{\textbf{Network}} & \textbf{Dataset} & \multicolumn{4}{c}{\textbf{CIFAR10}} & \multicolumn{4}{c}{\textbf{CIFAR100}} & \multicolumn{4}{c}{\textbf{TinyImageNet}} \\
 &  & \textbf{IPC} & DC & DSA & DM & MTT & DC & DSA & DM & MTT & DC & DSA & DM & MTT \\ \hline
\multirow{9}{*}{\textbf{Clean}} & \multirow{3}{*}{ResNet18} & 1 & 18.40 & 25.13 & 21.16 & 34.01 & 1.85 & 9.61 & 5.56 & 12.87 & 0.76 & 3.00 & 0.73 & 2.80 \\
 &  & 10 & 18.47 & 44.69 & 39.33 & 44.42 & 3.70 & 21.87 & 20.86 & 26.31 & 1.64 & 7.59 & 5.76 & 12.82 \\
 &  & 50 & 22.94 & 48.81 & 52.88 & 60.19 & 5.56 & 29.75 & 31.07 & 38.09 & 1.10 & 10.05 & 10.31 & 16.42 \\ \cline{2-15} 
 & \multirow{3}{*}{VGG11} & 1 & 26.50 & 23.50 & 22.71 & 28.32 & 9.62 & 10.75 & 6.65 & 15.31 & 1.11 & 5.34 & 3.38 & 7.85 \\
 &  & 10 & 35.38 & 44.11 & 40.89 & 47.92 & 12.96 & 21.49 & 21.40 & 27.62 & 4.22 & 9.92 & 10.44 & 16.96 \\
 &  & 50 & 37.70 & 49.38 & 54.65 & 61.75 & 6.51 & 28.70 & 27.97 & 34.69 & 0.67 & 8.72 & 9.52 & 15.05 \\ \cline{2-15} 
 & \multirow{3}{*}{MobileNet} & 1 & 15.14 & 20.63 & 14.48 & 16.88 & 1.26 & 7.65 & 1.67 & 3.37 & 0.63 & 2.79 & 0.67 & 2.55 \\
 &  & 10 & 10.78 & 28.08 & 29.31 & 27.62 & 2.41 & 13.22 & 11.46 & 11.80 & 0.93 & 5.03 & 4.51 & 6.47 \\
 &  & 50 & 15.66 & 33.68 & 35.98 & 32.67 & 4.74 & 9.01 & 9.17 & 9.78 & 0.81 & 3.18 & 2.28 & 2.17 \\ \hline
\multirow{9}{*}{\textbf{PGD}} & \multirow{3}{*}{ResNet18} & 1 & 0.00 & 14.50 & 10.58 & 11.08 & 0.00 & 3.80 & 1.93 & 2.09 & 0.00 & 0.12 & 0.00 & 0.01 \\
 &  & 10 & 0.00 & 8.02 & 10.68 & 5.37 & 0.00 & 4.33 & 6.45 & 2.72 & 0.00 & 0.13 & 0.45 & 0.10 \\
 &  & 50 & 0.00 & 3.05 & 7.39 & 7.08 & 0.01 & 2.21 & 2.34 & 0.95 & 0.00 & 0.03 & 0.89 & 0.00 \\ \cline{2-15} 
 & \multirow{3}{*}{VGG11} & 1 & 14.60 & 13.08 & 13.71 & 7.5 & 0.43 & 2.98 & 2.56 & 1.88 & 0.00 & 0.40 & 0.14 & 0.28 \\
 &  & 10 & 0.54 & 10.51 & 12.27 & 8.59 & 0.07 & 3.11 & 3.47 & 2.41 & 0.00 & 0.34 & 0.85 & 0.39 \\
 &  & 50 & 0.11 & 5.08 & 10.43 & 6.51 & 0.04 & 2.91 & 1.98 & 2.26 & 0.00 & 0.01 & 0.03 & 0.04 \\ \cline{2-15} 
 & \multirow{3}{*}{MobileNet} & 1 & 3.44 & 7.51 & 13.53 & 12.30 & 0.00 & 1.97 & 0.89 & 1.66 & 0.00 & 0.19 & 0.03 & 0.03 \\
 &  & 10 & 0.42 & 6.90 & 8.33 & 5.43 & 0.02 & 3.88 & 3.48 & 2.45 & 0.00 & 0.76 & 0.63 & 0.22 \\
 &  & 50 & 0.12 & 8.41 & 5.59 & 5.93 & 0.03 & 2.77 & 2.62 & 2.58 & 0.03 & 0.35 & 0.66 & 0.32 \\ \hline
\multirow{9}{*}{\textbf{CW}} & \multirow{3}{*}{ResNet18} & 1 & 0.00 & 14.01 & 9.63 & 10.09 & 0.00 & 3.01 & 1.35 & 1.05 & 0.00 & 0.00 & 0.00 & 0.00 \\
 &  & 10 & 0.00 & 7.81 & 10.14 & 4.98 & 0.00 & 3.96 & 5.74 & 2.58 & 0.00 & 0.09 & 0.44 & 0.10 \\
 &  & 50 & 0.00 & 3.26 & 7.81 & 7.58 & 0.00 & 2.55 & 2.66 & 1.12 & 0.03 & 0.09 & 0.76 & 0.00 \\ \cline{2-15} 
 & \multirow{3}{*}{VGG11} & 1 & 14.14 & 13.17 & 13.47 & 7.66 & 0.22 & 2.53 & 2.28 & 1.32 & 0.00 & 0.19 & 0.07 & 0.06 \\
 &  & 10 & 0.41 & 10.73 & 12.37 & 8.72 & 0.01 & 2.97 & 3.19 & 2.19 & 0.00 & 0.16 & 0.46 & 0.23 \\
 &  & 50 & 0.16 & 5.53 & 10.85 & 7.04 & 0.14 & 3.14 & 2.25 & 2.47 & 0.00 & 0.01 & 0.01 & 0.04 \\ \cline{2-15} 
 & \multirow{3}{*}{MobileNet} & 1 & 3.66 & 7.60 & 13.53 & 12.33 & 0.00 & 1.92 & 0.83 & 1.55 & 0.00 & 0.14 & 0.00 & 0.01 \\
 &  & 10 & 0.33 & 7.05 & 8.37 & 5.51 & 0.01 & 3.99 & 3.37 & 2.32 & 0.00 & 0.55 & 0.59 & 0.14 \\
 &  & 50 & 0.15 & 8.88 & 5.89 & 6.04 & 0.05 & 2.23 & 2.14 & 2.12 & 0.03 & 0.20 & 0.45 & 0.11 \\ \hline
\end{tabular}%
}
\end{table*}

\end{document}